%% file: iclr2026/iclr2026_conference.tex
\documentclass{article} % For LaTeX2e
\usepackage{iclr2026_conference,times}

\usepackage{caption}
\input{math_commands.tex}

\usepackage{hyperref}
\usepackage{url}
\usepackage[utf8]{inputenc} % allow utf-8 input
\usepackage[T1]{fontenc}    % use 8-bit T1 fonts
\usepackage{booktabs}       % professional-quality tables
\usepackage{amsfonts}       % blackboard math symbols
\usepackage{nicefrac}       % compact symbols for 1/2, etc.
\usepackage{microtype}      % microtypography
\usepackage{xcolor}         % colors

\usepackage{enumerate}
\usepackage[algo2e,ruled,noend,linesnumbered,norelsize]{algorithm2e}
\usepackage{ dsfont }
\usepackage{amsmath}
\usepackage{amsthm}
\usepackage{units}
\usepackage{mathtools}
\usepackage{color}
\usepackage{multirow}
\usepackage{enumitem}
\usepackage{booktabs}
\usepackage{array}
\usepackage{diagbox}

\usepackage{bm}
\usepackage{verbatim}
\usepackage{svg}
\usepackage{graphicx}
\usepackage{wrapfig}
\usepackage{adjustbox}
\usepackage{wrapfig}
\usepackage{amsmath}
\usepackage{ amssymb }

\newcommand\ourM{SCSF}

\newcommand{\rebuttal}[1]{\textcolor{black}{#1}}

\title{Accelerating Eigenvalue Dataset Generation \\via Chebyshev Subspace Filter}

\author{Antiquus S.~Hippocampus, Natalia Cerebro \& Amelie P. Amygdale \thanks{ Use footnote for providing further information
about author (webpage, alternative address)---\emph{not} for acknowledging
funding agencies.  Funding acknowledgements go at the end of the paper.} \\
Department of Computer Science\\
Cranberry-Lemon University\\
Pittsburgh, PA 15213, USA \\
\texttt{\{hippo,brain,jen\}@cs.cranberry-lemon.edu} \\
\And
Ji Q. Ren \& Yevgeny LeNet \\
Department of Computational Neuroscience \\
University of the Witwatersrand \\
Joburg, South Africa \\
\texttt{\{robot,net\}@wits.ac.za} \\
\AND
Coauthor \\
Affiliation \\
Address \\
\texttt{email}
}

\author{\textbf{Hong~Wang}\textsuperscript{1,2,3}\quad
\textbf{Jie~Wang}\textsuperscript{1,2,3}\thanks{Corresponding author.}\quad
\textbf{Jian~Luo}\textsuperscript{1,2,3}\quad
\textbf{Huanshuo~Dong}\textsuperscript{1,2,3}\quad
\textbf{Yeqiu~Chen}\textsuperscript{1,2,3}\quad \\
\textbf{Runmin~Jiang}\textsuperscript{1}\quad
\textbf{Zhen~Huang}\textsuperscript{1}
\\
\textsuperscript{1} University of Science and Technology of China\\
\textsuperscript{2} CAS Key Laboratory of Technology in GIPAS, University of Science and Technology of China\\
\textsuperscript{3} MoE Key Laboratory of Brain-inspired Intelligent Perception and Cognition, University of Sci-\\ence and Technology of China\\
{\small\tt wanghong1700@mail.ustc.edu.cn} \\
}

\iclrfinalcopy % Uncomment for camera-ready version, but NOT for submission.
\begin{document}

\maketitle

\begin{abstract}
Eigenvalue problems are among the most important topics in many scientific disciplines. With the recent surge and development of machine learning, neural eigenvalue methods have attracted significant attention as a forward pass of inference requires only a tiny fraction of the computation time compared to traditional solvers. 
However, a key limitation is the requirement for large amounts of labeled data in training, including operators and their eigenvalues.
To tackle this limitation, we propose a novel method, named \textbf{S}orting \textbf{C}hebyshev \textbf{S}ubspace \textbf{F}ilter (\textbf{SCSF}), which significantly accelerates eigenvalue data generation by leveraging similarities between operators---a factor overlooked by existing methods. 
Specifically, SCSF employs truncated fast Fourier transform sorting to group operators with similar eigenvalue distributions and constructs a Chebyshev subspace filter that leverages eigenpairs from previously solved problems to assist in solving subsequent ones, reducing redundant computations.
To the best of our knowledge, SCSF is the first method to accelerate eigenvalue data generation. 
Experimental results show that SCSF achieves up to a $3.5\times$ speedup compared to various numerical solvers.
\end{abstract}

\section{Introduction}\label{intro}

Solving eigenvalue problems is an important challenge in fields such as quantum physics \citep{pfau2023natural}, fluid dynamics \citep{schmid2010dynamic}, and structural mechanics \citep{wen2022u}.
Traditional numerical solvers, such as the Krylov-Schur algorithm \citep{stewart2002Krylov}, often suffer from prohibitively high computational costs when tackling complex problems.
% JL: Traditional numerical solvers like the Krylov-Schur algorithm \citep{stewart2002Krylov} are expensive for complex problems.
% \begin{figure}[H]
% \vskip 0in
% \begin{center}
% \begin{subfigure}
%     \centering
%     \raisebox{1cm}{\includegraphics[width=\linewidth]{figure/intro1_2_1.pdf}}
% \end{subfigure}
% \vskip -0.4in
% \begin{subfigure}
%     \centering
%     \includegraphics[width=\linewidth]{figure/intro2_1.pdf}
% \end{subfigure}
% \vskip -0.1in
% \caption{\textbf{Above.} Generation process of the eigenvalue dataset: 1. Generate a set of random parameters. 2. Derive the corresponding operators based on these parameters. 3. Convert the operators into matrices using discretization methods. 4. Independently solve for the matrix eigenvalues using numerical solvers. 5. Obtain the matrix eigenpairs, converting them into the operator eigenpairs. 6. Assemble the dataset. \textbf{Below.} Comparison of average computation times across various algorithms based on the number of eigenpairs solved.
% }
% \label{fg_intro}
% % \caption{\textbf{左图} Operator eigenvalue 数据集的生成过程。1. 生成一组随机参数。2. 根据这些参数导出相应的 Operator。3. 使用离散化方法将 Operator 转换为矩阵。4. 调用矩阵特征值求解器独立求解。5. 获取矩阵特征值、特征向量并将其转换为 Operator 特征值、特征函数。6. 组装成数据集。 \textbf{右图} 不同算法的平均计算时间和特征值求解个数的关系。从中可以看出，{\ourM} 算法可以显著加速特征值问题的求解。}
% \end{center}
% \vskip -0.25in
% \end{figure}
To overcome these computational challenges, recent advancements in deep learning \citep{schutt2017quantum, li2020fourier, luoneural} have demonstrated remarkable success as one forward pass only necessitates a tiny fraction of the computation time compared to numerical solvers, often in milliseconds.

Despite their success, data-driven approaches face a fundamental limitation: the reliance on labeled datasets. Training neural networks requires large-scale labeled data, which is often generated using computationally expensive traditional methods. 
It usually takes dozens of hours or even days. For example, the QM9 dataset \citep{ramakrishnan2014quantum} contains $1.34 \times 10^5$ molecular data points, each produced by solving Hamiltonian operator eigenvalue problems. These calculations typically employ traditional algorithms, whose computational costs can escalate dramatically with increasing problem complexity, like finer grid resolutions or higher accuracy requirements. This scalability issue represents a significant bottleneck for generating the labeled data needed to train deep learning models. Furthermore, the diversity of scientific problems leads to the need for a unique dataset for each scenario, which further intensifies this challenge of computational intractability. As a result, the high computational expense of generating eigenvalue data severely limits the application of data-driven approaches \citep{ren2026foundation,zhang2023artificial,huang2025self,liupapm,liu2024deep,liu2025aerogto,wu2025mpg, luo2024neural,liu2025efficient, zhang2024learning,wangmixture,hou2026learning,zhang2026hgatsolver}.

\begin{figure}[t]
  \centering
\hspace{-0.5cm}
  % 第一张图片
  \begin{minipage}{0.55\textwidth}
    \centering
    \includegraphics[width=1\linewidth]{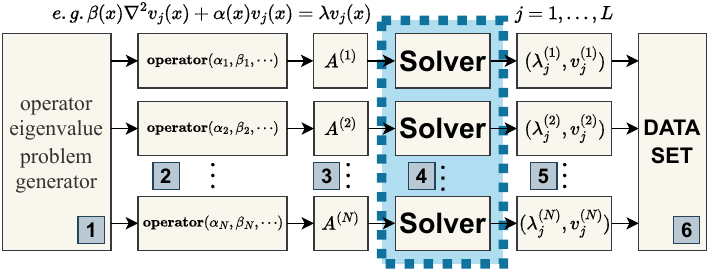}
  \end{minipage}
  % 第二张图片
  \begin{minipage}{0.45\textwidth}
    \centering
    \includegraphics[width=1\linewidth]{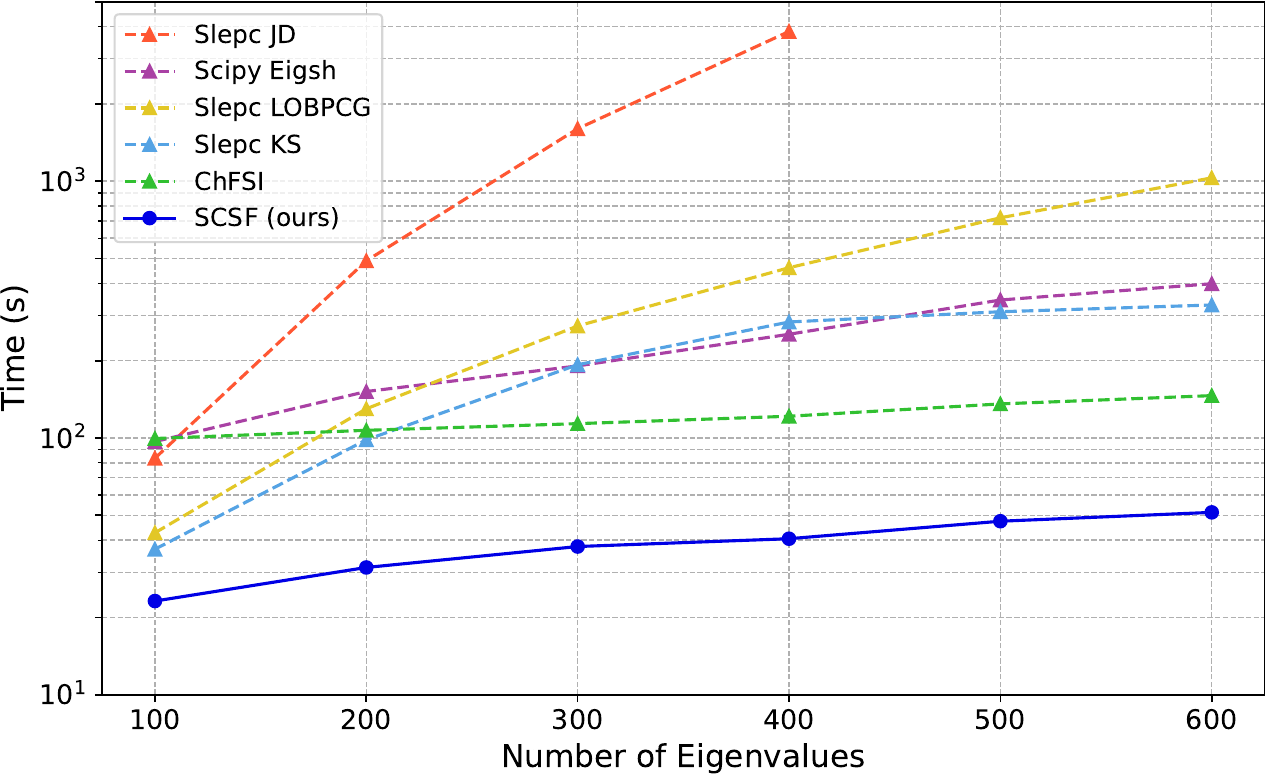}
  \end{minipage}
  \caption{\textbf{Left.} Generation process of the eigenvalue dataset: 1. Generate a set of random problem parameters. 2. Derive the corresponding operators based on these parameters. 3. Convert the operators into matrices using discretization methods. 4. Independently solve for the matrix eigenvalues using numerical solvers. 5. Obtain the matrix eigenpairs, converting them into the operator eigenpairs. 6. Assemble the dataset.
  \textbf{Right.} Results of average computation times across various algorithms based on the number of eigenvalues solved on the Helmholtz operator dataset.
  }
  \vspace{-0.5cm}
  % The variation in accuracy across iterations and time for NeurKItt compared to the baseline GMRES algorithm is illustrated. Notably, the NeurKItt algorithm substantially enhances the efficiency of solving the linear systems, with a reduction in the number of iterations by up to a factor of 13 and achieving a speed-up of up to 2 times.
  \label{fig:images}
\end{figure}

In particular, the dataset generation process typically involves six key steps, as illustrated in Figure~\ref{fig:images} (left). Among these steps, solving the eigenvalue problem is the most computationally demanding (step 4), accounting for 95$\%$ of the total processing cost~\citep{hughes2012finite}. Existing data generation methods typically compute the eigenvalues of each matrix in the dataset independently. 
However, operators in the dataset often share similarities, as they describe related physical phenomena, which can largely simplify and accelerate the eigenvalue-solving process. Existing approaches, however, fail to leverage these similarities, leading to significant computational redundancy.
% The generation of the eigenvalue dataset is illustrated in Figure~\ref{fg_intro_1}. 
% Notably, the computational cost in step 4 often accounting for over 95\% of the entire process~\citep{hughes2012finite}.
% % , underscoring it as a prime candidate for acceleration. 
% In step 4, existing data generation methods solve these problems independently. 
% The matrix eigenvalue computation phase (step 4) dominates the computational cost, consuming over 95\% of the total processing time~\citep{hughes2012finite}. Current methods address this computationally intensive step by solving each eigenvalue problem independently.
% In reality, problems with close parameters exhibit significant similarities, and solving them independently introduces redundant computations. 
%In fact, we argue that this similarity can be leveraged to accelerate the solution of matrix eigenvalue problems. 
Previous works \citep{wang2024accelerating, dong2024accelerating, liu2025accelerating} have demonstrated the potential of leveraging similarity to significantly reduce generation time of linear system datasets. However, how to effectively exploit matrix similarity to accelerate eigenvalue datasets generating remains an unknown problem.

% \vspace{-0.5cm}

% Generate eigenvalue problems by random parameters
%%%% WG
To address this problem, we introduce a novel data generation approach, named \textbf{S}orting \textbf{C}hebyshev \textbf{S}ubspace \textbf{F}ilter (\textbf{\ourM}).
% SCSF被设计成 利用相似问题的近似特征对，减少特征值求解过程的冗余计算，来加速数据集生成
{\ourM} is designed to use the eigenpairs of similar problems to reduce redundant computations in the eigenvalue solving process, thereby accelerating eigenvalue dataset generation.
Specifically, at the beginning, {\ourM} employs a sorting algorithm based on truncated Fast Fourier transform (FFT), which arranges these problems efficiently, enhancing the adjacent correlation between problems in the queue and laying the groundwork for sequential solving. 
Then, {\ourM} accelerates the convergence of iterations and significantly reduces computation times by constructing a Chebyshev subspace filter, which solves the problem aided by the eigenpairs from previous problem solving.
The core design of {\ourM} is to identify and exploit the close spectral distributions and invariant subspaces within these eigenvalue problems. 
% And {\ourM} does not treat each system as a discrete entity but rather cleverly coordinates their sequential resolution, eliminating substantial computational redundancy caused by similar structures.
{\ourM} coordinates the sequential resolution of these systems rather than treating them as discrete entities.
This improved approach not only alleviates the computational demands of the eigenvalue algorithm but also significantly speeds up the generation of training data for data-driven algorithms. %“creation”改成“generation”呢
% Extensive experimental results demonstrate {\ourM} successfully accelerates the operator dataset generation, achieving a 
We summarize our contributions as follows:
\begin{itemize}
    \item
    % We propose an approach to accelerate Krylov iterative methods, similar to warm-starting, by predicting invariant subspaces, speeding up the solution of general large-scale linear equation systems.
    % To the best of our knowledge, our work represents \textit{the first} attempt to apply data-driven approaches to optimize the Krylov subspace algorithms for solving generic non-symmetric linear systems.
    % To the best of our knowledge, {\ourM} is the first method to accelerate the operator eigenvalue data generation.
    \rebuttal{To the best of our knowledge, {\ourM} is the first method to accelerate operator eigenvalue dataset generation by exploiting intrinsic operator similarity through a novel sorting and filtering framework.}
    % 我们提出通过预测不变子空间的方式来加速Krylov迭代算法，从而极大地加速了一般大型线性方程组的求解
    % 我们的方法可以被结合给任意
    \item
    By using truncated FFT sorting and the Chebyshev filtered subspace iteration, we introduce a novel approach that transforms dataset generation into sequence eigenvalue problems.
    % We introduce a novel strategy that predicts the invariant subspace of the linear system to accelerate Krylov iteration. To facilitate the subspace prediction, we design a projection loss for efficient training, in conjunction with QR decomposition for stable outputs.
    % \item
    % \huanshuodong{We have introduced a neural operator for predicting the invariant subspace of linear systems to enhance iteration efficiency. This neural operator utilizes a specially designed projection loss based on QR decomposition to effectively facilitate the subspace prediction.}
    % We introduce a novel strategy, which predicts the invariant subspace of linear systems to significantly enhance iteration efficiency. We further design a projection loss, in conjunction with QR decomposition, to effectively accomplish this predictive task.
    % Through meticulously designed loss functions and QR decomposition, this network effectively predicts subspaces.
    % To efficiently implement the aforementioned invariant subspace prediction, we design the projection loss function for this task.
    % 我们为这个任务设计了专用的loss
    % 没有设计新的网络
    % 为了高效实现上述的不变子空间预测，我们设计了一种新的网络结构预测，loss设计
    \item%这句话可能要改一下
    % Extensive experiments demonstrate that {\ourM} significantly reduces the computational cost for eigenvalue dataset. As shown in Figure \ref{fg_intro} (below), compared to existing fastest method, {\ourM} achieves up to a $6.4 \times$ speedup.
    Comprehensive experiments demonstrate that {\ourM} substantially reduces the computational cost of eigenvalue dataset generation. As demonstrated in Figure \ref{fig:images} (right), our method achieves up to a $3.5\times$ speedup compared to state-of-the-art solvers.
    % 大量实验和理论分析表明，{\ourM} 降低了算子特征值问题数据集的计算成本。如图1所示，现有的最快方法相比，速度提高了高达 6.4 倍。
    %
\end{itemize}

\section{Preliminaries}
% 1. Discretization of PDEs in Neural Operator Training
% 2. 介绍不用filter谱变换的算法基座

\subsection{Discretization of Eigenvalue Problem}
\label{sec:discrete}
% 我们主要关注生成训练数据过程中时间占比最大的求解矩阵特征值问题部分。As illustrated in Figure~\ref{fg_intro}，通常会采用离散化数值方法求解它们，例如FDM、FEM、FVM~\citep{strikwerda2004finite, hughes2012finite, johnson2012numerical, leveque2002finite, cheng2023solving}。这些离散化数值算法将 PDE 问题从无限维希尔伯特函数空间嵌入到合适的有限维空间中，从而将 算子 问题转换为矩阵问题。我们提供了一个简单的例子来阐明所讨论的过程。生成线性方程组的详细过程可在附录~\ref{FDM example} 中找到。具体来说，我们讨论使用 FDM 求解二维泊松算子特征值问题，将其转换为矩阵特征值问题：
% Our focus primarily lies in the component of generating training data that consumes the most time, namely the solution of matrix eigenvalue problems. %这什么，太抽象了，gpt写这么花哨
% Our main focus is on solving the matrix eigenvalue problem, the most time-consuming part of eigenvalue data generation.
\rebuttal{This section outlines the discretization of operator eigenvalue problems into matrix form. Our method focuses on solving these matrix problems, which is the most time-consuming step in generating the required eigenvalue datasets.}
% 我们主要关注矩阵特征值求解，这是数据生成中计算开销最大部分
As shown in Figure~\ref{fig:images} (left), these problems are typically solved by numerical discretization methods such as FDM~\citep{strikwerda2004finite, leveque2002finite}. These discretization techniques embed the infinite-dimensional Hilbert space of operators into an appropriate finite-dimensional space, thereby transforming operator eigenvalue problems into matrix eigenvalue problems. We provide a simple example to clarify the discussed processes.
A detailed process can be found in Appendix~\ref{FDM example}. Specifically, we discuss the case that uses FDM to solve the eigenvalue problem of the two-dimensional Poisson operator, transforming it into a matrix eigenvalue problem:
\begin{equation}
\displaystyle k(x, y) \nabla^2 u(x, y) = \lambda u(x,y).
\end{equation}
We map the problem onto a \(2 \times 2\) grid (i.e., \( N_x = N_y = 2 \) and \( \Delta x = \Delta y \)), where both the variable \(u_{i,j}\) and the coefficients \(k_{i,j}\) follow a row-major order. This setup facilitates the derivation of the matrix eigenvalue equation:
% \begin{tiny}
\begin{equation}
\begin{bmatrix}
k_{1,1} & 0 & 0 & 0 \\
0 & k_{1,2} & 0 & 0 \\
0 & 0 & k_{2,1} & 0 \\
0 & 0 & 0 & k_{2,2}
\end{bmatrix}
\begin{bmatrix}
-4 & 1 & 1 & 0 \\
1 & -4 & 0 & 1 \\
1 & 0 & -4 & 1 \\
0 & 1 & 1 & -4
\end{bmatrix}
\begin{bmatrix}
u_{1,1} \\
u_{1,2} \\
u_{2,1} \\
u_{2,2}
\end{bmatrix}
=
\lambda
\begin{bmatrix}
u_{1,1} \\
u_{1,2} \\
u_{2,1} \\
u_{2,2}
\end{bmatrix}
.
\label{eq_p}
\end{equation}
% \end{tiny}
By employing various methods to generate the parameter matrices \( P = 
\begin{bmatrix}
k_{11} & k_{12}  \\
k_{21} & k_{22}  
\end{bmatrix}
. \) 
% \begin{equation}
% P = 
% \begin{bmatrix}
% k_{11} & k_{12}  \\
% k_{21} & k_{22}  
% \end{bmatrix}
% .
% \end{equation}
Such as utilizing Gaussian random fields (GRF) or truncated polynomials, we can derive Poisson operators characterized by distinct parameters.

Typically, training a neural network requires a number of data from $10^3$ to $10^5$~\citep{lu2019deeponet}. 
Such a multitude of eigenvalue systems, derived from the same distribution of operators, naturally \rebuttal{exhibit a high similarity} \citep{soodhalter2020survey}. 
% 相关的数据驱动工作本质上就是在学习这种相似性
% Related data-driven approaches essentially aim to learn this similarity~\citep{luoneural}.
It is precisely this similarity that is key to the effective acceleration of {\ourM}. We can conceptualize this as the task of solving a sequential series of matrix eigenvalue problems:
\begin{equation}
	 A^{(i)} v^{(i)}_j=\lambda^{(i)}_j v^{(i)}_j, \quad j=1,\cdots,L; \quad i=1,2,\cdots,N
	\label{eq:eq_basic}
\end{equation}
where $L$ is the number of eigenvalues to be solved, 
$N$ is the number of eigenvalue problems,
the matrix $ A^{(i)}\in \mathds{C}^{n \times n}$, the eigenvector $ v^{(i)}_j\in \mathds{C}^{n}$,and the eigenvalue $\lambda^{(i)}_j \in \mathds{C}$ vary depending on the operator. 
We define the eigenpairs as \( (\Lambda^{(i)}, V^{(i)}) \), with \( \Lambda^{(i)} = \text{diag}(\lambda_1^{(i)}, \dots, \lambda_{L}^{(i)}) \), \( V^{(i)} = [v^{(i)}_1 | \cdots | v^{(i)}_L] \), and $ |\lambda_1^{(i)}| \leqslant |\lambda_2^{(i)}| \dots \leqslant |\lambda_{L}^{(i)}|$.

\subsection{The Chebyshev Polynomials and Chebyshev Filter}\label{pr:cheb}
\begin{algorithm2e}[b]
    \caption{Chebyshev Filter~\citep{berljafa2015optimized}}
    \label{alg:ChebFilter}
    \KwIn{Matrix $A \in \mathbb{C}^{n \times n}$, vectors $Y_0 \in \mathbb{C}^{n \times k}$, degree $m \in \mathbb{N}$, and parameters $\lambda, c, e \in \mathbb{R}$.}
    \KwOut{Filtered vectors $Y_m = C_m(Y_0)$, where each vector $Y_{m,j}$ is filtered with a Chebyshev polynomial of degree $m$.}   
    $A = (A - cI_n)/e, \quad \sigma_1 = e / (\lambda - c)$\;
    $Y_1 = \sigma_1 A Y_0$\;
    % $s = \argmin_{j=1,\dots,k} \, m_j \neq 1$\; 
    % $s = \underset{{j=1,\dots,k}}{\text{argmin}} \, m_j \neq 1$\; 
    \For{$i = 1, \dots, m - 1$}{
        $\sigma_{i+1} = 1 / (2 / \sigma_1 - \sigma_i)$\;
        $Y_{i+1,1:m-1} = Y_{i,1:m-1}, \quad Y_{i+1,m:k} = 2 \sigma_{i+1} A Y_{i,m:k} - \sigma_{i+1} \sigma_i Y_{i,m:k}$\;
    }    
\end{algorithm2e}

\rebuttal{
This section details the Chebyshev filter, which is constructed using Chebyshev polynomials. Distinguished by their optimal uniform approximation properties and efficient three-term recurrence relations, these are among the most widely utilized orthogonal polynomials~\citep{mason2002Chebyshev, rivlin2020Chebyshev}.
% This section details the Chebyshev filter, which is built upon the stronger approximation properties of Chebyshev polynomials compared to other orthogonal polynomials~\citep{mason2002Chebyshev, rivlin2020Chebyshev}. 
This filtering technique is particularly effective for our method because it can efficiently construct a polynomial that isolates a target spectral interval, allowing us to use the solution of one eigenvalue problem to significantly accelerate the next.}
% Chebyshev filtered subspace iteration 和 Chebyshev 正交多项式有很大关系，下面来简单介绍下其基础概念。
% Chebyshev filtered subspace iteration is closely related to Chebyshev orthogonal polynomials~\citep{mason2002Chebyshev, rivlin2020Chebyshev}.
% ; we briefly introduce their fundamental concepts below~\citep{mason2002Chebyshev, rivlin2020Chebyshev}.
% % 切比雪夫多项式因为较强的逼近能力得到了广泛应用，The Chebyshev polynomials \(C_m(t)\) of degree \(m\) can be defined on the interval \([-1, 1]\), their expression to 
% \begin{equation}
% C_m(t) = \cos(m \cos^{-1}(t)), \quad |t| \leq 1.
% \end{equation}
% Chebyshev polynomials are widely used due to their strong approximation capabilities. 

The Chebyshev polynomials \( C_m(t) \) of degree \( m \) are defined on \([-1, 1]\) and are expressed as
\begin{equation}
C_m(t) = \cos(m \cos^{-1}(t)), \quad |t| \leq 1.
\end{equation}
% C_m(t)通常被叫做第一类切比雪夫多项式，满足如下递推关系:
\(C_m(t)\) commonly referred to as the Chebyshev polynomial of the first kind, satisfies the following recurrence relation:
\begin{equation}
C_{m+1}(t) = 2 t C_m(t) - C_{m-1}(t).
\end{equation}

For a Hermitian matrix \(A \in \mathbb{C}^{n \times n}\) and vectors \(Y_0 \in \mathbb{C}^{n \times k}\), we use the three-term recurrence relation that defines Chebyshev polynomials in vector form:
\begin{equation}
C_{m+1}(Y_0) = 2 A C_m(Y_0) - C_{m-1}(Y_0), \quad
C_m(Y_0) \equiv C_m(A) Y_0.
\end{equation}
The computation of \(C_m(Y_0)\) and the Chebyshev filter is described in Algorithm~\ref{alg:ChebFilter}. Let \(A'\) denote the previously solved related matrix, with \((\lambda_i', v_i')\) in ascending order, and \(\{\lambda_2', \dots, \lambda_L'\} \in [\alpha, \beta]\).
In Algorithm~\ref{alg:ChebFilter}, the parameter \(\lambda\) is typically approximated by \(\lambda_1'\), while \(c = \frac{\alpha + \beta}{2}\) and \(e = \frac{\beta - \alpha}{2}\) represent the center and half-width of the interval \([\alpha, \beta]\), providing estimates for the spectral distribution of \(A\).
% The computational procedure below is adapted from \cite{berljafa2015optimized}.

% Algorithm 2 generalizes the mechanism just described earlier to the case of an arbitrary number \(k\) of vectors \(Y\) approximating the eigenvectors corresponding to the values \(\{\lambda_1, \dots, \lambda_k\} \notin [\alpha, \beta]\). The values for \(\beta\) and \(\alpha\) are respectively estimated by the Lanczos step of Algorithm 1 and by \(\lambda_{1}^{(\ell - 1)}\), while \(\lambda_{1}^{(\ell - 1)}\) gives an estimate for \(\gamma\). The actual polynomial \(p_m(t)\) is not computed explicitly. What is calculated is its action on the vectors \(Y\) by exploiting the three-term recurrence relation that can also be used to define Chebyshev polynomials
% \[
% C_{m+1}(Y) = 2 H C_m(Y) - C_{m-1}(Y), \quad C_m(Y) \equiv C_m(H) \cdot Y.
% \]
% Equation (5) generalizes easily to \(p_m(Y)\) as shown in line 8 of Algorithm 2.

\section{Method}\label{method}
% 1. 总起 算法流程图 以及 伪代码细节
% 2. 排序
% 3. 切比雪夫 filter
% 4. deflation & locking subspace
\begin{figure*}[t]
    \centering
    \includegraphics[width=1\columnwidth]{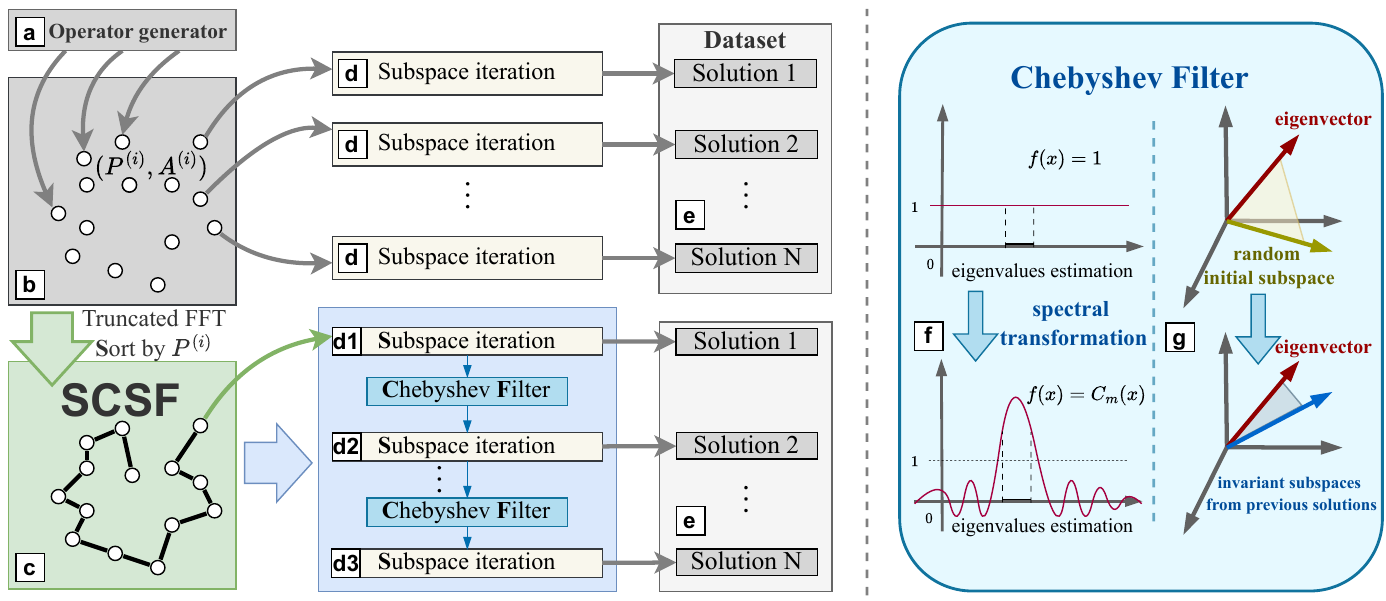}
    \caption{Algorithm Flow Diagram: \textbf{a}. Generation of operators to be solved. \textbf{b}. Discretization of operators into matrices. \textbf{c}. Apply {\ourM} algorithm to sort matrices, obtaining a sequence with strong correlations. \textbf{d}. Other algorithms independently solve eigenvalue problems. \textbf{d1, d2, d3}. {\ourM} algorithm utilizes Chebyshev subspace iterations to sequentially solve the eigenvalue problems. \textbf{e}. Assembly of eigenvalue pairs into a dataset. \textbf{f}. Amplification of the interval of interest through spectral transformation.  \textbf{g}. Replacement of initial subspaces with previously solved invariant subspaces.}
    \label{fg_method1}
    \vskip -0.10in
\end{figure*}

% \vspace{-0.37cm}
% 算法流程图：a.生成需求解的算子 b.将算子离散成矩阵 c.应用{\ourM}对矩阵进行排序,获得具有强相关性的序列 d.先前算法分别独立求解步骤b中的特征值问题 d1,d2,d3.{\ourM} 利用切比雪夫子空间迭代按顺序求解 e.获得 特征值对 并将其组装成数据集 f.通过谱变换放大感兴趣的区间 g.利用先前求解的不变子空间取代随机初始子空间

% 如图~\ref{fg_method1}所示，我们旨在通过利用这些特征值问题之间的内在相关性来加速解决方案。具体来说，当连续特征值问题表现出显著相关性时，可能是由于轻微的扰动，先前解决方案中的不变子空间和谱分布可能未得到充分利用。这些未开发的信息可能会加快后续特征值问题的解决速度。
% 我们利用已知的不变子空间和谱分布设计对应的Chebyshev Filtered Subspace Iteration，使得加速求解。这类算法在一些序列特征值问题中已经得到充分认可。

% JL: 这一章节我们将介绍xxx。xxxx主要是利用样本之间的相似性来加速求解的，它由两部分组成：排序和求解。简单来说，排序算法借助矩阵间的谱相似性来重新排列求解顺序，而求解算法则可以更好的继承谱性质接近的矩阵的求解结果，从而加速加速当前矩阵的求解。图xx所示为xx算法的流程图。
% JL: 首先在secxx我们介绍xx，并进一步在sec xx介绍xx。
% \jluo{This section introduces SCSF, a fast data generation approach that improves the efficiency of solving eigenvalue problems by leveraging intrinsic spectral correlations among operators. SCSF incorporates two key components: (1) a truncated Fast Fourier Transform (FFT)-based approach for efficiently sorting operator eigenvalue samples and (2) a strategy to ensure that successive systems in the sequence exhibit meaningful spectral correlations, thereby facilitating the application of Chebyshev filtered subspace iteration (ChFSI). By integrating these components, {\ourM} enables effective utilization of prior spectral information, accelerating the solution process for subsequent eigenvalue problems.}

In this section, we introduce our novel method, named the sorting Chebyshev subspace filter (SCSF), a fast data generation approach that efficiently solves eigenvalue problems by leveraging intrinsic spectral correlations among operators. SCSF incorporates two key components: (1) a truncated fast Fourier transform (FFT)-based approach for efficiently sorting operator eigenvalue problems and (2) the Chebyshev filtered subspace iteration (ChFSI) employed for sequential solving. By integrating these components, SCSF can use spectral information from the previous eigenvalue problem solving to aid the next eigenvalue problem solving, thus accelerating the eigenvalue data generation.

We first introduce the sorting algorithm that leverages the spectral similarities and provides the time complexity analysis in Section \ref{sec:sorting} . Then we give an introduction to the Chebyshev filtered subspace iteration in Section \ref{sec:ChFSI}. Figure \ref{fg_method1} shows the overview of our SCSF. Generally, the truncated FFT sorting algorithm ensures that successive matrices in the sequence exhibit close relations. Then ordered sequence enables ChFSI to effectively utilize prior information, thereby accelerating the solution process~\citep{berljafa2015optimized}.

\subsection{The Sorting Algorithm}
\label{sec:sorting}

To benefit the successive solving sequence of the eigenvalue problem, we need a sorting algorithm that pulls matrices with similar spectral properties, like invariant subspaces, close enough in the solving sequence, so that solving the current matrix in sequence can be easily boosted by the previous solving. 
% Recalling Section \ref{sec:discrete}, eigenvalue problem, the matrix $A^{(i)}$, is generated from the parameter matrix $P^{(i)}$~\citep{lu2022comprehensive, li2020fourier}.
\rebuttal{
Recalling Section \ref{sec:discrete}, the matrix $A^{(i)}$ is the operator matrix derived through numerical discretization (e.g., Finite Difference Method) from the parameter matrix $P^{(i)}$ (e.g., the coefficient $k(x,y)$ in the Poisson equation)~\citep{lu2022comprehensive, li2020fourier}. For instance, in the Poisson equation, the values $k_{i,j}$ from $P^{(i)}$ become part of the discretized sparse matrix $A^{(i)}$, as shown in Eq. \ref{eq_p}.
} 
A naive strategy is to use the Frobenius distance of the parameter matrices $P^{(i)}$ to perform a greedy sort~\citep{wang2024accelerating}.
And by repeatedly fetching without reservation from the remaining matrix in the dataset, we can reorganize the solving sequence so that the successive solving can benefit from the re-ordered sequence.
\begin{algorithm2e}[tb]
	\caption{The Truncated FFT Sorting Algorithm}
	\label{alg:sort}
	\SetAlgoLined
	\KwIn{Sequence of eigenvalue problems to be solved $ A^{(i)}\in \mathds{C}^{n \times n}$, corresponding parameter matrix $ P^{(i)}\in \mathds{C}^{p \times p}, i= 1,2,\cdots,N$, $p_0$ is the truncation threshold for low frequencies, and $ P^{(i)}_{low}\in \mathds{C}^{p_0 \times p_0}$.}
	\KwOut{Sequence for eigenvalue problems ${seq}_{mat}$.}
        Initialize the list with sequence $seq_0 = \{1,2,\cdots,N\}$, ${seq}_{mat}$ is an empty list\;
        Set $i_0 = 1$ as the starting point. Remove $1$ from $seq_0$ and append $1$ to ${seq}_{mat}$\;
    \For{$i=1, \cdots, N$}{
            Let $ P^{(i)}_{low} =  \text{Trunc}_{p_0}\left( \text{FFT}(P^{(i)}) \right)$. Perform truncated FFT on matrix $P^{(i)}$ to extract low-frequency information\;
    }
	\For{$i=1, \cdots, N-1$ and $dis = 1000$}{
            % $dis = 1000$\;
		\For{each $j$ in $seq_0$ }{
		$dis_j = $ the Frobenius norm of the difference between $P^{(i_0)}_{low}$ and $P^{(j)}_{low}$\;
            \If{$dis_j \textless dis$}{
                $dis = dis_j$ and $j_{min} = j$\;
            }
	}
        Remove $j_{min}$ from $seq_0$, append $j_{min}$ to ${seq}_{mat}$ and set $i_0 = j_{min}$\;
	}
        Get the sequence for eigenvalue problems ${seq}_{mat}$\;
\end{algorithm2e}
% 然而，由于排序算法过程中需要反复计算不同参数矩阵距离，这是其核心的计算开销，这和矩阵边长的大小即PDE问题的分辨率直接相关

% 相关研究表明，影响PDE问题的主要变量来自参数矩阵的低频部分，高频部分往往表示为PDE的噪声等无关数据。基于这点，为了减少排序的计算开销，我们在依据参数矩阵排序之前，先对矩阵进行截断FFT变换，提取PDE问题的低频信息，再通过对比低频信息之间的距离来排序。

However, the main computational cost of such a naive sorting algorithm arises from repeatedly calculating the distances between different matrices $P$, which is directly related to the matrix dimension—that is, the resolution of operators. 
% , which is discrete result of the given operator $u(x)$. 
Existing works \citep{holmes2012turbulence, li2020fourier} have shown that the key variables that affect operators stem from the low-frequency components of the parameter matrices $P$, while high-frequency components often represent noise or irrelevant data. 
Based on this insight, to reduce computational overhead during sorting, we first perform a truncated FFT on the parameter matrices to extract the low-frequency information before sorting. We then sort by comparing the Frobenius distances between these low-frequency components.

% 如算法~\ref{alg:sort}所示，不妨假设，特征值问题个数为N个，参数矩阵P的边长为p，低频截断矩阵的边长为k。那么直接使用贪心算法的计算复杂度为$\mathcal{O}(N^2p^2)$。我们的排序算法的计算复杂度有以下两部分组成：1.  FFT 的复杂度为 $\mathcal{O}(p^2 \log p)$，总共N个矩阵，这部分共$ \mathcal{O}(Np^2 \log p )$ 2.后续贪心排序算法 $\mathcal{O}(N^2k^2)$。总共$ \mathcal{O}(N^2k^2)+Np^2 \log p )$，而一般来说，k<<p，p<<N，由此可见我们的排序算法可以有效减少计算开销。
As shown in Algorithm~\ref{alg:sort}, suppose we have $N$ eigenvalue problems, the parameter matrices $P^{(i)}\in \mathds{C}^{p \times p}$, and the low-frequency truncated matrices $P^{(i)}_{low}\in \mathds{C}^{p_0 \times p_0}$.
The computational complexity of directly using a greedy algorithm is $\mathcal{O}(N^2p^2)$. Our sorting algorithm's complexity consists of two main parts: 1. {FFT Computation}: The complexity of FFT is $\mathcal{O}(p^2 \log p)$ per matrix. For $N$ matrices, this totals $\mathcal{O}(Np^2 \log p)$.
2. {Greedy Sorting}: The subsequent greedy sorting algorithm has a complexity of $\mathcal{O}(N^2p_0^2)$.
Overall, the total complexity is $\mathcal{O}(N^2p_0^2 + Np^2 \log p)$. Since $p_0 \ll p$ and $p \ll N$, our sorting algorithm effectively reduces computational cost.
\rebuttal{In addition, we conducted a qualitative theoretical analysis of the truncated FFT to demonstrate its rationality. Details can be found in Appendix~\ref{ap_fft_theory}.}

\subsection{Chebyshev Filtered Subspace Iteration}
\label{sec:ChFSI}

% % 对于一系列特征值问题，连续系统之间存在固有相关性。因此，我们假设通过利用先前系统求解过程中的信息，我们可以加快后续系统的迭代收敛，从而显著提高计算性能。 对于各种类型的 PDE 问题，生成的特征值问题表现出具有不同结构特征的矩阵。这些独特的矩阵结构与 Chebyshev filtered subspace iteration (ChFSI)相一致。为了验证我们算法的有效性，我们主要关注PDE算子特征值问题中最普遍和最普遍的场景，算子的自伴的，此时自伴算子对应的矩阵 \(A\) 是hermit的。
% In a series of eigenvalue problems, inherent correlations often exist between successive systems. 
% We hypothesize that leveraging the eigenpairs $(\Lambda^{(i-1)}, V^{(i-1)})$ of the previous problem $A^{(i-1)}$ can accelerate the iterative convergence of the subsequent system $A^{(i)}$, thereby significantly enhancing computational performance. 
% For various types of operators, the resulting eigenvalue problems produce matrices with distinct structural characteristics. These unique matrix structures align well with the ChFSI method~\citep{manteuffel1977tchebychev, saad2011numerical, winkelmann2019chase, berljafa2015optimized}. To verify the effectiveness of our algorithm, we focus on the most common scenario in eigenvalue problems where the operator is self-adjoint; in this case, the corresponding matrix \(A\) is Hermitian.

% 通过排序算法我们得到前后强相关的特征值问题序列。我们采用Chebyshev filtered subspace iteration (ChFSI)利用先前系统求解过程中的信息，加快后续系统的迭代收敛，从而显著提高计算性能。我们主要关注PDE算子特征值问题中最普遍和最普遍的场景，算子的自伴的，此时自伴算子对应的矩阵 \(A\) 是hermit的。

After the sorting algorithm, we obtain a sequence of eigenvalue problems that exhibit strong correlations between consecutive problems. We employ the Chebyshev filtered subspace iteration~\citep{manteuffel1977tchebychev, saad2011numerical, winkelmann2019chase, berljafa2015optimized} that leverages the eigenpairs $(\Lambda^{(i-1)}, V^{(i-1)})$ of the previous problem $A^{(i-1)}$ to accelerate the iterative convergence of the subsequent problem $A^{(i)}$, thereby significantly enhancing computational performance. 
We focus on the most common scenario in eigenvalue problems where the operator is self-adjoint; in this case, the corresponding matrix \(A\) is Hermitian.

% 算法~\ref{alg:ChFSI}展示了ChFSI求解第i个特征值问题的过程（i>1），其中初始近似的不变子空间$ V^{(i-1)}$和谱分布$\Lambda^{(i-1)}$来自于序列中前一个特征值问题的特征向量和特征值，m表示其中filter function的多项式度数，一般选取10到15.如果是序列中的第一个特征值问题(i=1)则随机生成初始迭代子空间\tilde{\Lambda}_0和初始谱\tilde{V}_0求解。
Algorithm~\ref{alg:ChFSI} outlines the process of ChFSI for solving the $i$-th eigenvalue problem $A^{(i)}$ ($1 < i \leqslant N$) where $L$ eigenvalues need to be solved.
The initial approximate invariant subspace $V^{(i-1)}$ and spectral distribution $\Lambda^{(i-1)}$ are derived from the eigenvectors and eigenvalues of the previous problem $A^{(i-1)}$ in the sequence. 
% The parameter $m$ denotes the polynomial degree in the filter function, e.g., $m = 20$. 
\rebuttal{
This initialization strategy, illustrated in Figure~\ref{fg_method1} (g), acts as a ``warm start'', where a high-quality initial subspace significantly reduces the computational cost of subsequent iterations. The parameter $m$ denotes the polynomial degree in the filter function, e.g., $m=20$. As depicted in Figure~\ref{fg_method1} (f), the Chebyshev filter modifies the operator's spectrum to amplify the target region, effectively transforming the problem into an equivalent one that is easier to solve. 
}
For the first eigenvalue problem $A^{(1)}$ in the sequence, the initial iterative subspace $\tilde{V}_0$ and initial spectrum $\tilde{\Lambda}_0$ are randomly generated.

% 具体来说ChFSI先通过少量 Lanczos 迭代（第 3 行）和已知的近似特征值来估计特征问题谱的上界 [31]，该估计值可以辅助后续filter function的构造。

% （第 5 行）通过的向量版本的切比雪夫多项式进行filter变换，详情可见Preliminaries。在切比雪夫滤波器步骤之后，跨越不变子空间的向量块很容易变得线性相关，一般采用正交化手段来避免这件事。在 ChFSI 中，该正交化程序是通过基于 Householder 反射器的 QR 分解（第 6 行）建立的。然后（第 7 行）使用分解的 \tilde{V}_0 来计算矩阵 A^{(i)} 的瑞利-里兹商。这样的商表示特征问题到逼近所寻求的特征空间的子空间的投影。（第 9 行）将所得的简化特征问题对角化，并将计算出的特征向量投影回更大的问题。在 Rayleigh-Ritz 步骤结束时，计算特征向量残差，锁定收敛的特征对，并将未收敛的向量设置为再次过滤。对于每个非收敛向量，（第 11 行）使用其相应的残差和近似特征值计算多项式滤波器的优化度。

% Specifically, ChFSI begins by estimating the upper bound of the eigenvalue spectrum (\texttt{line 2}) using a few Lanczos iterations and known approximate eigenvalues~\citep{zhou2011bounding, saad2011numerical}. 
% This estimate aids in constructing the subsequent filter function. 
In \texttt{line 3}, the Chebyshev filter is applied using the vector form of Chebyshev polynomials; details can be found in the preliminaries Section~\ref{pr:cheb}. 
After the Chebyshev filtering step, the vector block $\tilde{V}_0$ spanning the invariant subspace may become linearly dependent. 
To prevent this, orthonormalization is performed (\texttt{line 4}) using QR decomposition based on Householder reflectors. 
\texttt{Line 5} computes the Rayleigh quotient of $A^{(i)}$ using the orthonormalized $\tilde{V}_0$, projecting the eigenvalue problem onto a subspace that approximates the desired eigenspace. 
In \texttt{line 6}, the reduced eigenvalue problem is diagonalized, and the computed eigenvectors are projected back to the original problem. 
At the end of the Rayleigh-Ritz step, relative residuals of the computed eigenvectors are calculated; converged eigenpairs are locked, and non-converged vectors are set to be filtered again (\texttt{line 7}). 
% For each non-converged vector, the optimal degree of the polynomial filter is updated (\texttt{line 11}) based on its residual and approximate eigenvalue.
% In \texttt{line 9}, convergence is determined based on the relative residual.

% 不妨假设， $m$ the degree of the polynomial, $n$ the size of the eigenproblem, and $k$ the number of filtered vectors. 则该算法每一轮的迭代的计算复杂度有以下几部分组成：1.\textsc{Chebyshev Filter} is $\mathcal{O}(mn^2k)$, 2. the QR factorization accounts for $\mathcal{O}(nk^2)$,3. the whole \textsc{Rayleigh--Ritz} procedure amounts to $\mathcal{O}(n^2k + nk^2 + k^3)$, 4. the residuals check in the \textsc{Locking} step is $\mathcal{O}(n^2k)$。由于$m \gg 1$，filter 部分是其中最为占时间的部分。
\begin{algorithm2e}[t]
    \caption{Chebyshev Filtered Subspace Iteration}
    \label{alg:ChFSI}
    
    \KwIn{Eigenvalue problem $A^{(i)}$, eigenpairs $(\Lambda^{(i-1)}, V^{(i-1)})$ of the previous eigenvalue problem $A^{(i-1)}$ where $\Lambda^{(i-1)} = \text{diag}(\lambda_1^{(i-1)}, \dots, \lambda_{L}^{(i-1)})$, \( V^{(i-1)} = [v^{(i-1)}_1 | \cdots | v^{(i-1)}_L] \), and filter degree $m$.}
    \KwOut{Wanted eigenpairs $(\Lambda^{(i)}, V^{(i)})$.}
    
    Initialize empty arrays/matrices $(\tilde{\Lambda}, \tilde{V})$, and set $\tilde{\Lambda}_0 = \Lambda^{(i-1)}$, $\tilde{V}_0 = V^{(i-1)}$\;
    % Estimate the largest eigenvalue of matrix \( A \) via Lanczos iteration\;
    \Repeat{the number of converged eigenpairs $\geq$ L}{
        Apply Chebyshev filter:  $\tilde{V}_0  = C_m(\tilde{V}_0 )$\;
        Perform QR orthonormalization on $Q R = [\tilde{V} | \tilde{V}_0 ]$\;
        Compute Rayleigh quotient $G = Q^{\top}_0 A^{(i)}  Q $\;
        Solve the reduced problem $GW = W\tilde{\Lambda}_0$, and update  $\tilde{V}_0  = \tilde{V}_0 W$\;
        Lock converged eigenpairs into $(\tilde{\Lambda}, \tilde{V})$\;
    }
    Return eigenpairs $(\Lambda^{(i)}, V^{(i)}) = (\tilde{\Lambda}, \tilde{V})$ \;
\end{algorithm2e}
Assuming $m$ is the degree of the polynomial, $n$ is the dimension of the matrix $A$, and $L$ is the number of eigenvalues to be solved, the computational complexity per iteration comprises:
1. {Chebyshev filter}: $\mathcal{O}(mn^2L)$
2. {QR factorization}: $\mathcal{O}(nL^2)$
3. {Rayleigh-Ritz procedure}: $\mathcal{O}(n^2 L + nL^2 + L^3)$
4. {Residuals check}: $\mathcal{O}(n^2L)$
. Since $m \gg 1$ and $ n \gg L$, the Chebyshev filtering step is the most computationally intensive.

% Chebyshev filtered subspace iteration中的加速有效性在很大程度上取决于以促特征值问题后续迭代收敛的方式选择近似的不变子空间和特征值。正确的排序可以放大它们的影响，从而减少迭代次数。这强调了排序的重要性

% \begin{figure}[htb]
%     \centering
%     \includegraphics[width=6cm]{figure/exp2.pdf}
%     % \caption{Comparison of average computation times across different algorithms as a function of matrix dimension for the generalized Poisson operator dataset, solving for 400 eigenvalues with a precision of \(1e-12\).}
%     \caption{000}
%     % 在广义泊松算子数据集下,特征值求解个数为400个,求解精度为1e-12,对比不同算法随矩阵边长变化的平均计算时间.
%     % 平均计算时间和矩阵维度的关系图，在广义泊松算子数据集下,特征值求解个数为400个,求解精度为1e-12。
%     \label{fg_exp1}
% \end{figure}

The acceleration of the Chebyshev filtered subspace iteration heavily depends on selecting approximate invariant subspaces and eigenvalues that promote rapid convergence in subsequent iterations. Proper sorting amplifies their impact, reducing the number of iterations required. This underscores the critical importance of the sorting algorithm in our method.

\section{Experiment}\label{experiment}

\subsection{Experimental Settings}\label{expsetting}
% To comprehensively evaluate the performance of {\ourM} in comparison to another algorithm, we conducted nearly 3,000 experiments. The detailed data is available in the Appendix~\ref{exp data}. Each experiment utilized a data set crafted to emulate an authentic NO training data set. Our analysis centered on two primary performance metrics viewed through three perspectives. These tests spanned four different datasets, with SKR consistently delivering commendable results. Specifically, the three Perspectives are:
% 1. Matrix preconditioning techniques, spanning 7 to 10 standard methods.
% 2. Accuracy criteria for linear system solutions, emphasizing 5 to 8 distinct tolerances.
% 3. Different matrix sizes, considering 5 to 6 variations. 
% Our primary performance Metrics encompassed:
% 1. Average computational time overhead.
% 2. Mean iteration count.
% For a deeper dive into the specifics, please consult the Appendix~\ref{exp_main}.

% 为了全面评估 {\ourM} 与其他算法相比的性能，我们进行了广泛的实验,每次实验都模拟生成一个算子特征值数据集。我们主要对比不同的特征值求解个数和不同的矩阵大小下的平均计算时间。这些测试涵盖了三个不同的数据集和4种主流的特征值求解算法，{\ourM} 始终如一地提供值得称赞的结果。
% 如需深入了解具体信息，请参阅附录 ~\ref{exp_main}。
% related work 可见附录A

To comprehensively assess the performance of our approach {\ourM} against other algorithms, we conducted extensive experiments, each simulating the generation of an operator eigenvalue dataset. We primarily compared the average computation times across different numbers of eigenvalues solved and various matrix sizes. These tests encompassed four distinct datasets and five mainstream eigenvalue solving algorithms, with {\ourM} consistently delivering commendable results. 
The detailed data is provided in Appendix \ref{ap_exp_main}, and the related work is discussed in Appendix \ref{ap:related}.

% \textbf{基线。} 如前所述，我们的重点围绕自伴微分算子导出的矩阵特征值问题，它一般由大型稀疏hermit矩阵组成。我们以以下几种主流算法的专业库实现作为baseline: 1. Scipy库的Eigsh (Implicitly Restarted Lanczos Method ) 2.SLEPc库的LOBPCG算法 3.SLEPc库的Krylov-Schur算法 4.SLEPc库的Jacobi-Davidso算法.

\textbf{Baseline}. Our focus solves the eigenvalue problem of matrices derived from self-adjoint differential operators, typically consisting of large Hermitian matrices. We benchmarked against the following mainstream algorithms implemented in libraries widely used: 1. Eigsh from SciPy (implicitly restarted Lanczos method)~\citep{2020SciPy-NMeth}, 2. Locally optimal block preconditioned conjugate gradient (LOBPCG) algorithm from SLEPc~\citep{knyazev2001toward, str-6}, 3. Krylov-Schur (KS) algorithm from SLEPc~\citep{stewart2002Krylov}, 4. Jacobi-Davidson (JD) algorithm from SLEPc~\citep{sleijpen2000jacobi}, 5.  Chebyshev filtered subspace iteration (ChFSI)~\citep{berljafa2015optimized} with random initialization. For detailed information, please refer to Appendix \ref{exp_base}.

% \textbf{Datasets.}
% To probe the algorithm's adaptability across matrix types, we delved into four distinct linear equation challenges, each rooted in a PDE: 
% 1. Darcy Flow Problem~\citep{li2020fourier, rahman2022u, kovachki2021neural, lu2022comprehensive}; 2. Thermal Problem~\citep{sharma2018weakly, koric2023data}; 3. Poisson Equation~\citep{hsieh2019learning, zhang2022hybrid}; 4. Helmholtz Equation~\citep{zhang2022hybrid}. For an in-depth exposition of the dataset and its generation, kindly refer to the Appendix~\ref{data set}. For the runtime environment, refer to Appendix~\ref{exp_env}.

% \textbf{数据集。}
% 为了探究该算法在矩阵类型中的适应性，我们深入研究了三个不同的算子特征值问题挑战：
% 1.广义泊松算子 ;2. 二阶椭圆偏微分算子 3. 亥姆霍兹算子; 有关数据集及其生成的深入说明，请参阅附录~\ref{数据集}。所有实验均讨论矩阵模意义下最小L个特征值及其特征向量的求解,L是求解数量 (这对应算子的特征值和特征函数),有关运行时环境，请参阅附录~\ref{exp_env}。

\textbf{Datasets}.
To explore the adaptability of the algorithm across different matrix types, we investigate four distinct operator eigenvalue problems: 1. Generalized Poisson operator; 2. Second-order elliptic partial differential operator; 3. Helmholtz operator; 4. Fourth-order vibration equation. For a thorough description of the datasets and their generation, please refer to Appendix \ref{dataset}. 

All experiments focus on computing the smallest $L$ eigenvalues in absolute value and their corresponding eigenvectors. 
For the runtime environment, experimental parameters, and parallelism setup, please see the Appendixes~\ref{exp_env},~\ref{expparam}, and~\ref{sec:appendix_parallelism}.
The hyperparameter analysis experiments, runtimes for various components of {\ourM}, the reliability of data generated by traditional algorithms, can be found in Appendixex~\ref{hperp},~\ref{exp_time}, and~\ref{sec:appendix_ground_truth}.
% 

% 注:本文实验均采用相对残差作为求解精度的指标,其定义可见附录A. {\ourM}是一种数值代数算法,可以根据需要调整求解精度.{\ourM}是一种纯加速的技术,在指定精度下,不会改变求解的结果.本文实验精度均在1e-8以上,远高于常见的神经网络相对误差(1e-1到1e-5),可以视为ground true.因此,不同的数值算法得到的数据集不会影响神经网络训练的效果.
We note that all experiments use relative residual as the metric for solution precision, with its definition provided in Appendix \ref{ap:rel}. {\ourM} is a numerical algebra algorithm that allows for adjustable solution precision as needed. It is purely an acceleration technique and does not alter the solution results at the specified precision. 
The solution precision for all experiments is set to at least 1e-8, which is significantly higher than the typical relative error range of neural networks (1e-1 to 1e-5), making it effectively a ground truth. 
Therefore, the datasets generated by different numerical algorithms will not affect the training performance of neural networks.

\subsection{Main Experiment}
% 主实验

% \begin{table}[htb]
% \centering
% \caption{Comparison of average computation times (in seconds) for eigenvalue problems using various algorithms. The first row lists different algorithms, the first column details the datasets including matrix size and solution precision, and the second column shows the number of eigenvalues computed for each matrix. The best algorithm is in bold.The symbol `-' denotes data not recorded due to excessive computation times.}
% % 比较使用不同算法求解特征值问题的平均计算时间（秒）。第一行对应不同的算法,第一列列出了数据集以及其矩阵边长和求解精度,第二列列出了每个矩阵求解的特征值个数, 其中最佳方法以粗体显示。
% \label{tab:main1}
% \begin{center}
% % \begin{normalsize}
% \renewcommand{\arraystretch}{1.2}
% % \resizebox{0.7\textwidth}{!}{%
%         \scriptsize
% \begin{tabular}{lcc|cccccc}
% \toprule
% {Dataset} & {L} &&& {Eigsh} & {LOBPCG} & {KS} & {JD} & {{\ourM} (ours)} \\ 
% % \midrule
% \hline
% Poisson & 200 &&& 14.20 & 73.03 & 23.76 & 270.2 & \textbf{12.85} \\ 
% 2500 & 300 &&& 26.27 & 151.5 & 45.95 & 920.8 & \textbf{25.61} \\ 
% 1e-12 & 400 &&& 36.86 & 265.3 & 72.32 & 2691 & \textbf{33.91} \\ 
% \hline
% Ellipse & 200 &&& 41.82 & 139.2 & 61.77 & 414.3 & \textbf{24.08} \\ 
% 4900 & 300 &&& 62.47 & 264.1 & 110.5 & 1446 & \textbf{29.88} \\ 
% 1e-10 & 400 &&& 87.19 & 459.7 & 188.7 & 3386 & \textbf{34.60} \\ 
% \hline
% Helmholtz & 200 &&& 151.7 & 129.9 & 98.34 & 489.6 & \textbf{31.31} \\ 
% 6400 & 400 &&& 253.5 & 460.4 & 283.0 & 3829 & \textbf{40.52} \\ 
% 1e-8 & 600 &&& 398.8 & 1031 & 329.6 & - & \textbf{51.32} \\ 
% \bottomrule
% \end{tabular}
% % }
% % \end{normalsize}
% \renewcommand{\arraystretch}{1}
% \end{center}
% % \vskip -0.1in
% \end{table}

\begin{table}[t]
\vskip -0.1in
\centering
\caption{Comparison of average computation times (in seconds) for eigenvalue problems using various algorithms. The first row lists different algorithms, the first column details the datasets, including matrix dimensions and solution precisions (relative residual), and the second column shows the number of eigenvalues $L$ computed for each matrix. The best algorithm is in \textbf{bold}. The symbol `-' denotes the result of a method that fails to converge under the given setting.}
% 比较使用不同算法求解特征值问题的平均计算时间（秒）。第一行对应不同的算法,第一列列出了数据集以及其矩阵边长和求解精度,第二列列出了每个矩阵求解的特征值个数, 其中最佳方法以粗体显示。
\label{tab:main1}
\begin{center}
% \begin{normalsize}
\renewcommand{\arraystretch}{0.8}
\resizebox{1\textwidth}{!}{%
        \tiny
\begin{tabular}{ccc|ccccccc}
\toprule
{Dataset} & $L$ &&& {Eigsh} & {LOBPCG} & {KS} & {JD} & {ChFSI}& {SCSF (ours)} \\ 
% \midrule
\midrule
Poisson & 200 &&& 14.20 & 73.03 & 23.76 & 270.2& 24.00 & \textbf{12.85} \\ 
2500 & 300 &&& 26.27 & 151.5 & 45.95 & 920.8& 38.03 & \textbf{25.61} \\ 
1e-12 & 400 &&& 36.86 & 265.3 & 72.32 & 2691& 57.41 & \textbf{33.91} \\ 
\midrule
Ellipse & 200 &&& 41.82 & 139.2 & 61.77 & 414.3& 43.90 & \textbf{24.08} \\ 
4900 & 300 &&& 62.47 & 264.1 & 110.5 & 1446& 60.69 & \textbf{29.88} \\ 
1e-10 & 400 &&& 87.19 & 459.7 & 188.7 & 3386& 67.13 & \textbf{34.60} \\ 
\midrule
Helmholtz & 200 &&& 151.7 & 129.9 & 98.34 & 489.6& 107.1 & \textbf{31.31} \\ 
6400 & 400 &&& 253.5 & 460.4 & 283.0 & 3829& 121.5 & \textbf{40.52} \\ 
1e-8 & 600 &&& 398.8 & 1031 & 329.6 & - & 146.2& \textbf{51.32} \\ 
\midrule
Vibration & 200 &&& 397.9 & 333.7 & 272.0 & 1230& 300.8 & \textbf{85.70} \\ 
10000 & 400 &&& 635.6 & 1170 & 768.8 & -& 310.5 & \textbf{107.2} \\ 
1e-8 & 600 &&& 1037 & 2716 & 857.8 & - & 382.3& \textbf{131.4} \\ 
\bottomrule
\end{tabular}
}
% \end{normalsize}
\renewcommand{\arraystretch}{1}
\end{center}
\vskip -0.13in
\end{table}

Table~\ref{tab:main1} showcases selected experimental data. From this table, we can infer several conclusions:
First, across all settings, our {\ourM} consistently has the lowest computation cost. 
The most significant improvements appeared in the Helmholtz dataset, where SCSF demonstrated speedups of $8\times$, $20\times$, $6\times$, $95\times$, and $3.5\times$ compared to Eigsh, LOBPCG, KS, JD, and ChFSI algorithms, respectively. These results confirm that SCSF effectively reduces inherent redundancies in sequential eigenvalue problems, substantially accelerating operator eigenvalue dataset generation.

% 其次, 随着每个矩阵求解特征值个数$num$的增加,相比其他算法{\ourM}算法的求解速度变得更加明显.例如,在二阶椭圆算子数据集上,在求解200个特征值时,{\ourM}是Krylov-Schur算法2.5倍,在求解400个特征值时,求解速度达到了5.5倍. 
% 这是因为对于Krylov-Schur算法来说,求解要从一维开始生成子空间.随着$num$的增加,需要更多的迭代次数,这极大增大了计算开销. 而我们的{\ourM}从先前的求解结果中继承近似的不变子空间,以此为基础开始迭代,扩大了初始的搜索空间以及可以利用的信息。{\ourM}随着$num$的增加,迭代次数的变化并不明显,使得计算开销的变化也不明显.

\begin{wrapfigure}{r}{0.47\textwidth} % r 表示图片在右侧，0.5\textwidth 表示图片占页面宽度的一半
% \vskip 0.2in
    \centering
    \includegraphics[width=\linewidth]{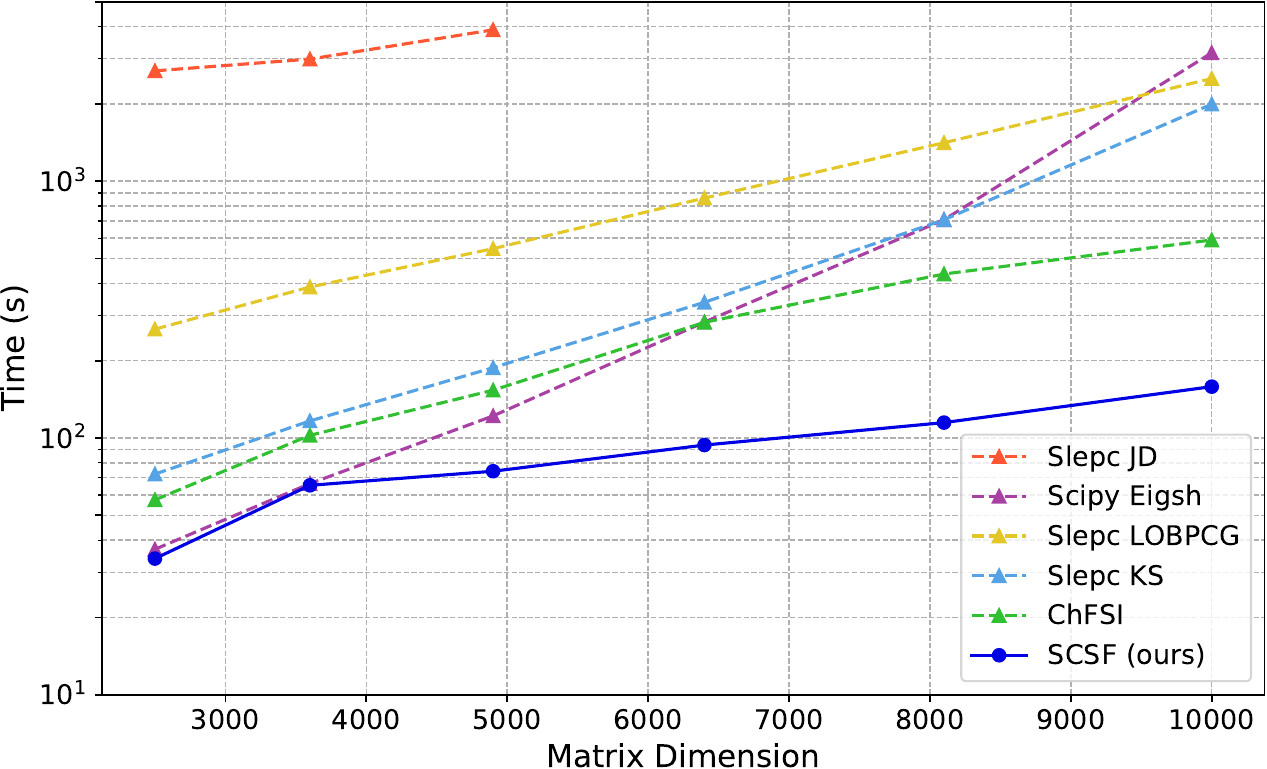}
    \caption{Plot of average computation time versus matrix dimension for solving 400 eigenvalues with a precision of 1e-12 on the generalized Poisson operator dataset.}
    \label{fg_exp1}
    \vskip -0.1in
\end{wrapfigure}

Moreover, as the number of eigenvalues \(L\) solved per matrix increases, the speed advantage of {\ourM} over other algorithms becomes more pronounced. 
For instance, on the second-order elliptic operator dataset, when solving for 200 eigenvalues, {\ourM} is 2.5 times faster than the Krylov-Schur method and 5.5 times faster at 400 eigenvalues. 
This efficiency stems from {\ourM} inheriting approximate invariant subspaces from previous solutions, effectively leveraging available information to expand the initial search space. Consequently, SCSF requires minimal additional iterations as $L$ increases, resulting in modest computation time growth.

% This efficiency stems from {\ourM} inheriting approximate invariant subspaces from previous solutions, thus expanding the initial search space and leveraging available information more effectively. 
% As \(L\) increases, the variation in {\ourM}'s iteration count is minimal, which in turn minimizes changes in computational overhead.
% As a result, the number of iterations required by {\ourM} does not change significantly with the increase in $L$, leading to a minimal increase in computation time.
% 这导致随着$num$的增加,SCSF所需迭代次数变化并不明显,使得计算时间不怎么变化.

% 第三, 不同数据集上求解效果差距很大.例如,在广义泊松算子数据集上,{\ourM}计算速度只略领先Scipy的Eigsh 10%左右,然而在Helmholtz数据集上,领先4-7倍. 一方面,不同算子的数值性质和矩阵组装格式都会直接影响到算法的效果.另一方面,很有可能是矩阵边长对其的效果的影响.为了分析矩阵边长的影响,我们做了补充实验,具体实验数据可见附录D.如图3所示,矩阵边长越大{\ourM}的效果越明显,在矩阵边长小于3600时,{\ourM}和Eigsh的计算效率相近.在矩阵边长大于5000,{\ourM}明显优于Eigsh和其他算法.

Besides, the performance disparity across different datasets is significant. For example, on the generalized Poisson operator dataset, {\ourM} is only about 10\% faster than Eigsh, yet it leads by 4-7 times on the Helmholtz dataset. 
This difference can be attributed to the numerical properties of different operators and the matrix assembly formats, which directly influence algorithmic performance.

We also conducted additional experiments to show that the impact of the matrix dimension is also significant. Results are shown in Figure~\ref{fg_exp1}, {\ourM} performs noticeably better as matrix dimensions increase. Below the matrix dimension of 3600, {\ourM} and Eigsh show comparable efficiency. However, beyond 5000, {\ourM} significantly outperforms Eigsh and other algorithms. For more details about matrix dimension influence, we refer to the results in Appendix~\ref{exp_size}.

This phenomenon can be explained through operator matrix approximation. A fixed operator has invariant eigenvalues and eigenfunctions. Varying matrix dimensions correspond to embedding the operator in different finite-dimensional linear spaces. For a fixed number of eigenvalues \(L\), larger matrices yield more accurate approximations of the true eigenvalues. That is, larger matrices reduce computational noise and enhance operator similarity visibility, enabling {\ourM} to utilize these similarities more effectively for superior performance. For comparisons with neural networks, similarity impact, and edge-case performance, see Appendices~\ref{sec:appendix_unsupervised},~\ref{sec:appendix_bounds}, and~\ref{sec:appendix_failure_case}.

\subsection{Efficacy of Chebyshev Subspace Filter}
\begin{table*}[htb]
\centering
\caption{Impact of initial subspace modifications on average computation time (in seconds) for different algorithms. `*' denotes the modified version. The first row lists algorithms, and the first column shows the number of eigenvalues \( L \) computed. The best algorithm is in \textbf{bold}, and `-' indicates the result of a method that fails to converge under the given setting.}
\label{tab:ab1}
\begin{center}
% \begin{normalsize}
\renewcommand{\arraystretch}{1.1}
\resizebox{1\textwidth}{!}{%
        \small
\begin{tabular}{l|ccccccccc}
\toprule
{$L$} & {Eigsh} & {Eigsh*} & {LOBPCG} & {LOBPCG*} & {KS} & {KS*} & {JD} & {JD*} & {SCSF (ours)} \\ 
\midrule
% \hline
200 & 151.7 & 150.2 & 129.9 & 95.9 & 98.34 & 100.6 & 489.6 & 760.1 & \textbf{31.31} \\ 
300 & 208.8 & 206.3 & 270.1 & 199.8 & 179.9 & 185.2 & 1803  & 3101  & \textbf{38.67} \\
400 & 253.5 & 249.1 & 460.4 & 362.1 & 283.0 & 292.2 & 3829 & 6374 & \textbf{40.52} \\ 
500 & 324.6 & 315.3 & 717.3 & 573.7 & 314.2 & 317.4 &  -  &   -  &  \textbf{46.70} \\
600 & 398.8 & 394.7 & 1031  & 866.0 & 329.6 & 335.7 &  -  &  -  &  \textbf{51.32} \\
\bottomrule
\end{tabular}
}
% \end{normalsize}
\renewcommand{\arraystretch}{1}
\end{center}
\vskip -0.1in
\end{table*}

% 为了分析Chebyshev Subspace Filter的有效性，我们进行了以下实验。在我们排序后，将现有算法的初始向量或子空间均设置为前一个问题的特征向量。对比计算时间。在Helmholtz 算子数据集上进行测试，矩阵维度6400，求解精度1e-8。结果如表2所示，*表示修改初始子空间的版本。
To analyze the efficacy of the Chebyshev subspace filter, we conducted the following experiments. 
After sorting, the initial vector or subspace for the existing algorithms was set to the eigenvectors from the previous problem (the modified version `*'). 
We compared the computational time across different methods. All experiments were conducted on the Helmholtz operator dataset, with a matrix dimension of 6400 and a tolerance of 1e-8. The results are shown in Table~\ref{tab:ab1}.
% , where * indicates the version with the modified initial subspace.

% 首先，所有实验中{\ourM}的计算时间均为最小。这充分说明了Chebyshev Subspace Filter的有效性。同时说明Chebyshev Subspace Filter是利用问题相似信息来降低冗余的最佳选择。
% 其次，通过修改初始设置会对不同算法的影响不同。1.lobpcg有明显加速。其底层逻辑和{\ourM}相似，都是通过对子空间不断优化来求解。初始子空间对求解影响很大。2.Eigsh、KS几乎没有影响。它们是从一个初始向量开始，通过Krylov迭代来求解的。也就是说，初始向量最多只能影响到单个特征值的求解，对整体时间影响不大。3.JD的性能有所下降。因为其性能和初始子空间的大小有关。我们的操作改变了默认初始子空间的维数。
First, the computation time for {\ourM} in all experiments was minimal, clearly demonstrating the efficacy of the Chebyshev subspace filter. This also highlights that the Chebyshev subspace filter is the optimal choice for leveraging problem similarity to reduce redundancy.

% Second, modifying the initial setup had varying impacts on different algorithms.
% 1. LOBPCG: showed significant acceleration. Its underlying logic is similar to {\ourM}, both relying on iterative optimization of the subspace to solve the problem. The initial subspace has a considerable impact on the solution.
% 2. Eigsh and KS were almost unaffected. These methods start with an initial vector (rather than a subspace) and solve the problem through Krylov iteration. 
% In other words, problem similarity only impacts a vector, with little effect on the overall time.
% 3. JD showed a performance decline. This is because its performance is sensitive to the size of the initial subspace. Our modification altered the default dimension of the initial subspace.
Second, the impact of initial setup modifications varied across algorithms:
1. LOBPCG accelerated significantly due to its subspace-based logic, similar to {\ourM}, where initialization strongly influences convergence.
2. Eigsh and KS remained largely unaffected as they rely on initial vectors and Krylov iteration, making problem similarity less impactful.
% 3. JD performed worse due to its sensitivity to the size of the initial subspace, which was altered in our modification.
3. JD showed a performance decline. This is because its performance is sensitive to the size of the initial subspace. Our modification altered the default dimension of the initial subspace.

\begin{table}[htb]
\centering
\caption{Performance comparison of {\ourM} with and without sorting. The first column lists the number of eigenvalues $L$ computed, while subsequent columns display average computation times, average iteration counts, total Flop counts, and filter Flop counts. Experiments used the matrix dimension of 2500 and precision 1e-12 on the generalized Poisson operator dataset.}
% 使用和不使用排序的{\ourM}性能比较，第一列为求解的特征值个数，各列分别展示求解的平均计算时间、平均迭代次数、总Flop次数、Filter部分的Flop次数。实验在广义泊松算子数据集，矩阵边长2500，求解精度1e-12下进行。
\label{tab:sort1}
\begin{center}
% \begin{normalsize}
\renewcommand{\arraystretch}{1.1}
\resizebox{1\textwidth}{!}{%
        \small
 \begin{tabular}{l|cc|cc|cc|cc}
\toprule
\multirow{2}{*}{$L$} & \multicolumn{2}{c|}{{Time (s)}} & \multicolumn{2}{c|}{{Iteration}} & \multicolumn{2}{c|}{{Flops}} & \multicolumn{2}{c}{{Filter Flops}} \\ 
& {\; w/o sort \;} & {\; sort \;} & {\; w/o sort \;} & {\; sort \;} & {\; w/o sort \;} & {\; sort \; }  & {\; w/o sort \;} & {\; sort \; }  \\ 
% \midrule
\midrule
20 & 8.248 & 2.971 & 19.70 & 9.880 & 519.7 & 298.4 & 485.8 & 280.8 \\ 
100 & 14.18 & 9.891 & 18.77 & 15.38 & 1984 & 1332 & 1798 & 970.1 \\ 
200 & 18.45 & 12.85 & 36.30 & 33.67 & 4459 & 3944 & 3654 & 3192 \\ 
300 & 34.59 & 25.61 & 47.50 & 39.18 & 8967 & 7544 & 6985 & 5702 \\ 
400 & 42.60 & 33.91 & 47.43 & 45.18 & 12022 & 11182 & 9087 & 8338 \\ 
\bottomrule
\end{tabular}
}
% \end{normalsize}
\renewcommand{\arraystretch}{1}
\end{center}
\vskip -0.15in
\end{table}

\subsection{Efficacy of Sorting Algorithms}

We analyze the performance of the sorting algorithm module from two perspectives: 
1. Comparing the performance of {\ourM} algorithm with and without `sorting' as shown in Table~\ref{tab:sort1}. 
2. Evaluating the effectiveness of different sorting algorithms as detailed in Tables~\ref{tab:sort2} and~\ref{tab:sort3}.
% 注:不排序的{\ourM}近似于直接使用Chebyshev subspace filter.不同于主实验中的ChFSI,不排序的{\ourM}每次求解的初始化设置为前一个问题求解得到的信息(未经排序的默认序列).
We note that if the setting is `w/o sort', {\ourM} is approximately equivalent to directly using the Chebyshev subspace filter. Unlike the ChFSI used in the main experiments, the initialization of each solve in the `w/o sort' {\ourM} is set based on the information obtained from solving the previous problem (following the default unsorted sequence).

\begin{wraptable}{r}{0.48\textwidth} % r 表示右侧，0.5\textwidth 表示宽度占页面的一半
\centering
\renewcommand{\arraystretch}{1}
\small
\caption{Comparison of average computation times (in seconds) for different sorting algorithms, with the first column indicating dataset size. Experiments used the matrix dimension of 6400 on the Helmholtz dataset.}
\begin{tabular}{l|c|ccc}
\toprule
\multirow{2}{*}{{Size}}   & \;Greedy\; &  \multicolumn{3}{c}{\;\;\;Truncated FFT Sort (ours)\;\;}\\ 
 & Total &  \;\;FFT & Greedy & Total \\ 
\midrule
$10^2$ & 0.114 &  \;0.0016 & 0.0147 & 0.0163 \\ 
$10^3$ & 7.328 &  \;0.0164 & 1.421 & 1.438 \\ 
$10^4$ & 592.7 &  \;0.1658 & 150.9 & 151.1 \\ 
\bottomrule
\end{tabular}
\renewcommand{\arraystretch}{1}
% \vskip -0.1in
\label{tab:sort2}
\end{wraptable}

Firstly, Table~\ref{tab:sort1} indicates that incorporating sorting can improve {\ourM} speed to 1.3 to 2.8 times, reduce the number of iterations by 5\% to 50\%, and decrease total Flops by 7\% to 43\%. 
The effect of sorting is more pronounced with smaller numbers of solutions $L$. This is because when $L$ is large, the inherited subspace already contains most of the necessary correlation information, diminishing the impact of sorting. Moreover, the Flops in the Filter component constitute over 70\% of {\ourM}'s computational load. A detailed time analysis of different aspects of {\ourM} can be found in Appendix~\ref{exp_time}.
% 此外,The `w/o sort' {\ourM}比主实验中的ChFSI计算速度提升到1.2-1.5times.它们的主要区别是初始化不一样.ChFSI每次求解均采用随机初始化,而The `w/o sort' {\ourM}采用前一个问题的信息进行初始化.这说明就算不经过排序,数据集中的问题之间存在一定相似性.这种相似性可以有效加速求解.
Additionally, the `w/o sort' {\ourM} achieves a computational speedup of 1.2 to 1.5 times compared to the ChFSI used in the main experiments. The primary difference lies in their initialization strategies: ChFSI uses random initialization for each solve, whereas the `w/o sort' {\ourM} leverages information from the previous problem for initialization. 
This indicates that, even without sorting, there is a certain level of similarity between problems in the dataset. Such similarity can effectively accelerate the solving process.

Secondly, as shown in Table~\ref{tab:sort2}, our designed truncated FFT sorting algorithm incurs significantly lower time cost compared to the complete greedy sorting in SKR~\citep{wang2024accelerating}, with its benefits becoming more pronounced as the dataset size increases.
In the truncated FFT sorting algorithm, the FFT contributes minimally to computational overhead but significantly reduces the time required for subsequent greedy sorting. 
% Table~\ref{tab:sort3} shows {\ourM} solution times for matrices sorted using either greedy or truncated FFT sorting are nearly identical, highlighting its effectiveness.

\newpage

\begin{wraptable}{r}{0.4\textwidth} % r 表示右侧，0.5\textwidth 表示宽度占页面的一半
    \centering
    % \vskip -0.1in
    \renewcommand{\arraystretch}{1.1}
    \small
    \caption{Comparison of average computation times and iteration counts for different sorting algorithms using {\ourM}. Experiments used the matrix dimension of 6400 on the Helmholtz dataset, precision 1e-8, and targeting 400 eigenvalues.}
    \begin{tabular}{l|ccc}
    \toprule
    & {w/o sort} & {Greedy} & {Ours} \\ \midrule
    Time (s) & 66.66 & 40.52 & 40.52 \\ 
    Iteration & 10.4 & 5.5 & 5.5 \\ 
    \bottomrule
    \end{tabular}
    \renewcommand{\arraystretch}{1}
    \label{tab:sort3}
\end{wraptable} 
Table~\ref{tab:sort3} validates the accuracy of our efficient sorting against the expensive greedy approach. To ensure fairness, we averaged the results over 100 independent runs to eliminate fluctuations. The results for `Greedy' and `Ours' are identical because the sequences produced by both algorithms are over 98\% identical. 
Consequently, the downstream solver incurs exactly the same operation cost, demonstrating that our low-cost sorting achieves the same acceleration performance as the high-cost greedy approach.

%  此外,经我们实验只要截断FFT选取合理的参数(例如截断到k=20,k远小于参数矩阵边长P)就会有很好的效果,无需选取很大.相关实验可见附录A
\rebuttal{Furthermore, our experiments demonstrate that the choice of the truncation threshold, $p_0$, is robust. We found that excellent performance is achieved with a relatively small $p_0$ (e.g., $p_0=20$), a value much smaller than the parameter matrix dimension $p$. This is because performance stabilizes once a sufficient number of low-frequency components are captured, making it unnecessary to retain a large number of components. For related experiments, please refer to Appendix~\ref{ap:trun}.}
% Furthermore, our experiments show that as long as the truncated FFT is configured with reasonable parameters (e.g., truncating at \(p_0=20\), where \(p_0\) is much smaller than the dimension \(p\) of the parameter matrix \(P\)), it achieves excellent performance without the need for a large \(p_0\). For related experiments, please refer to Appendix~\ref{ap:trun}.

%  排序算法参数对最终性能的影响的补充实验可见附录A

\section{Conclusions and Future Work}\label{limit}
\paragraph{Conclusions}
In this paper, we addressed the critical bottleneck of generating large-scale eigenvalue datasets for training neural operators. 
We introduced SCSF, the first method to accelerate eigenvalue dataset generation by exploiting operator similarity. By integrating a truncated FFT sorting algorithm with a Chebyshev subspace filter, SCSF transforms the generation task into an efficient sequential solving problem. Our method achieves up to a 3.5$\times$ speedup over traditional solvers, significantly reducing computational redundancy. 
By lowering a key barrier to entry, SCSF provides a valuable tool for advancing research in the AI for Science community.

\paragraph{\rebuttal{Limitations and Future Work}}

\rebuttal{
Despite SCSF's demonstrated efficiency, we have identified three key directions for future development. First, we aim to extend the current framework from Hermitian linear operators to non-Hermitian and non-linear operators. Second, the correlation between operators is currently measured using the Frobenius norm of their parameter matrices. We plan to explore more effective, problem-specific metrics rather than relying on a fixed one. Finally, SCSF's effectiveness hinges on a continuous mapping between parameters and operators, rendering it ineffective when this mapping is discontinuous. Future work will involve measuring correlation directly from the operators or their matrix representations to bypass the challenges of discontinuous parameterization.
}

\section*{Acknowledgements}
% The authors would like to thank all the anonymous reviewers for their insightful comments.
% This work was supported in part by National Key R\&D Program of China under contract 2022ZD0119801, National Nature Science Foundations of China grants U23A20388, U19B2026, U19B2044, and 62021001.
% This research is supported by Smart-Grid National Science and Technology Major Project (Grant No.2025ZD0805500), National Key R\&D Program of China (Grant No. 2022ZD0119801), National Nature Science Foundations of China grants 124B1019, U23A20388, U19B2026, U19B2044, and 62021001.
The authors would like to thank all the anonymous reviewers for their insightful comments and valuable suggestions. 
% This work was supported by Smart-Grid National Science and Technology Major Project under contract 2025ZD0805500, the National Key R\&D Program of China under contract 2022ZD0119801, and the National Nature Science Foundations of China grants 124B1019, U23A20388, 62021001 and 624B1011.
This work was supported by the National Key R\&D Program of China (No. 2022ZD0119801), the National Natural Science Foundation of China (Nos. U23A20388, 62021001), and the Smart-Grid National Science and Technology Major Project (No. 2025ZD0805500).

\newpage

\bibliography{iclr2026_conference}
\bibliographystyle{iclr2026_conference}

%%%%%%%%%%%%%%%%%%%%%%%%%%%%%%%%%%%%%%%%%%%%%%%%%%%%%%%%%%%%
\newpage
\appendix

\section{Usage of LLMs}
\label{sec:llm_usage}

Throughout the preparation of this manuscript, Large Language Models (LLMs) were utilized as a writing and editing tool. Specifically, we employed LLMs to improve the clarity and readability of the text, refine sentence structures, and correct grammatical errors. All final content, including the core scientific claims, experimental design, and conclusions, was conceived and written by us, and we take full responsibility for the final version of this paper.

% 1. erting PDEs to Linear Systems: An Example
% 2. Algorithmic Details
%   2. 谱变换的具体细节
%   3. 子空间利用的具体细节

\section{Related work}\label{ap:related}

\subsection{Eigenvalue Datasets and Neural Eigenvalue Methods}
Eigenvalue datasets are widely utilized in neural eigenvalue methods.
% 在分子化学研究中，特征值算法通常用于确定关键的分子特征，例如轨道能级~\citep{kittel2018introduction}。这些特征构成了数据集的基础，是通过求解薛定谔方程和哈密顿算子的特征值获得的~\citep{helgaker2013molecular}。
% 该领域的著名数据集包括 QM7~\citep{blum}、QM9~\citep{ramakrishnan2014quantum}、ANI-1~\citep{smith2017ani} 和 MD17~\citep{chmiela2017machine}。
In molecular chemistry research, eigenvalue algorithms are commonly employed to determine critical molecular properties, such as orbital energy levels~\citep{kittel2018introduction}. These properties form the foundation of datasets and are obtained by solving the eigenvalue problem of the Schrödinger equation and the Hamiltonian operator~\citep{helgaker2013molecular}. Prominent datasets in this domain include QM7~\citep{blum}, QM9~\citep{ramakrishnan2014quantum}, ANI-1~\citep{smith2017ani}, and MD17~\citep{chmiela2017machine}.
% 在材料研究中，特征值算法通常用于求解材料的电子能态结构和态密度。相关的数据集例如Materials Project和OQMD。
In materials science, eigenvalue algorithms are often applied to solve for electronic band structures and density of states in materials. Representative datasets in this field include the materials project~\citep{jain2020materials} and OQMD~\citep{kirklin2015open}.
% 这些数据集已被广泛用于训练和验证神经特征值方法~\citep{schutt2017quantum, bartok2017machine, rupp2012fast}，从而推动了分子性质预测和材料设计领域的研究。
These datasets have been extensively used to train and validate neural eigenvalue methods~\citep{schutt2017quantum, bartok2017machine, rupp2012fast}, driving advancements in molecular property prediction and materials design.
% 在流体动力学和结构力学中，特征值算法往往用于模态分析。
% 近期出现了许多数据驱动的模态分析算法，它们需要对应微分算子的特征值数据集进行训练。
In fluid dynamics and structural mechanics, eigenvalue algorithms are frequently utilized for modal analysis. Recently, many data-driven modal analysis algorithms have emerged, requiring eigenvalue datasets corresponding to differential operators for training~\citep{murata2020nonlinear, iwata2023neural, alford2022deep, brunton2020machine, akkad2023dynamic, wangstnet}.
% 此外，一些工作需要算子特征值数据集来优化算法。例如\cite{luoneural}通过预测算子的特征函数来加速线性方程的求解。
Additionally, some studies leverage operator eigenvalue datasets to optimize algorithms. For instance, \cite{luoneural} accelerates the solution of linear systems by predicting the eigenfunctions of operators.

\subsection{Eigenvalue Data Generation Algorithms}

Training data-driven algorithms require a large amount of labeled eigenvalue data. Typically, the generation of these high-precision data is obtained by traditional algorithms.
% 计算数学的领域，对于算子PDE特征值问题，复杂PDE的数值求解通常涉及使用各种离散化方法~\citep{morton2005numerical} 将算子的特征值转换为可解的矩阵特征值问题，例如有限差分法 (FDM)~\citep{strikwerda2004finite}、有限元法 (FEM)~\citep{hughes2012finite, johnson2012numerical} 和有限体积法 (FVM)~\citep{leveque2002finite}。这些方法最终会形成大型矩阵特征值问题, 随着计算数学的发展，已经开发出各种特征值算法~\citep{golub2013matrix, watkins2007matrix}。幂法和逆迭代法~\citep{parlett1973geometric} 、雅可比法~\citep{lotkin1956characteristic}、 QR 算法~\citep{francis1962qr} 适用于求解较小矩阵的特征值问题。
% 在处理大型稀疏矩阵时，Lanczos 迭代~\citep{lanczos1950iteration} 和 Arnoldi 迭代~\citep{arnoldi1951principle} 分别为对称和非对称矩阵的特征值问题提供了定制解决方案，通过迭代构建 Krylov 子空间来提高计算效率。局部最优块预条件共轭梯度 (LOBPCG) 算法~\citep{knyazev2001toward}、Krylov-Schur 算法~\citep{stewart2002Krylov} 特别适用于计算大规模稀疏对称矩阵中的多个特征值和特征向量。
% Rayleigh-Ritz 方法适用于需要从大规模系统中精确提取部分特征值和特征向量的场景，该方法通过构建有效的子空间并在此基础上求解更小规模的矩阵问题，从而实现高效且精确的计算~\citep{golub2013matrix}。
In the field of computational mathematics, solving operator eigenvalue problems often involves utilizing various discretization methods such as finite difference methods (FDM)~\citep{strikwerda2004finite}, finite element methods (FEM)~\citep{hughes2012finite, johnson2012numerical}. 
These discretization methods transform operator eigenvalue problems into matrix eigenvalue problems, which are then solved using the corresponding matrix algorithms.
% For smaller matrices, common algorithms include the power method \citep{parlett1973geometric}, Jacobi method \citep{lotkin1956characteristic}, and QR algorithm \citep{francis1962qr}. 
For larger matrices, the Krylov-Schur algorithm \citep{stewart2002Krylov}, Jacobi-Davidson \citep{sleijpen2000jacobi}, and locally optimal block preconditioned conjugate gradient (LOBPCG) \citep{knyazev2001toward} are among the most frequently employed algorithms~\citep{golub2013matrix}.
% For large sparse matrices, Lanczos~\citep{lanczos1950iteration} and Arnoldi iterations~\citep{arnoldi1951principle} customize solutions via Krylov subspace constructions. The LOBPCG~\citep{knyazev2001toward} and Krylov-Schur algorithms~\citep{stewart2002Krylov} excel in computing multiple eigenvalues and vectors in large sparse symmetric matrices. Additionally, the Rayleigh-Ritz method (RR) efficiently computes specific eigenvalues and vectors from large-scale systems by solving reduced matrix problems within effectively constructed subspaces~\citep{golub2013matrix}.
% 对于大型矩阵来说，Krylov-Schur 算法~\citep{stewart2002Krylov},Jacobi-Davidson 和 LOBPCG~\citep{knyazev2001toward} 是最为常见的算法。

% 擅长计算大型稀疏对称矩阵中的多个特征值和向量。此外，瑞利-里兹方法 (RR) 通过在有效构造的子空间内求解简化矩阵问题，有效地计算大规模系统的特定特征值和向量~\citep{golub2013matrix}。

Nonetheless, traditional methods were not designed for dataset generation, resulting in high computational costs, which have become a significant barrier to the advancement of data-driven approaches~\citep{zhang2023artificial, hao2022physics, chen2025deepcontour}. 
Recent data augmentation research~\citep{brandstetter2022clifford, liu2023ino} has led to the development of methods that preserve symmetries and conservation laws, enhancing model generalization and data efficiency.
\cite{wang2024accelerating, dong2024accelerating} report acceleration in the process of solving linear equations, thereby speeding up the generation of PDE datasets. 

However, these improvements largely focus on neural networks or the rapid solution of linear system-based PDEs, without discussing optimizations in the generation of eigenvalue datasets.
% \cite{wang2024accelerating, dong2024accelerating}通过加速了线性方程组求解过程，从而加速了PDE数据集的生成。然而，这些进步主要集中在 PDE 求解算法本身的优化和线性方程组类PDE的快速求解，并未讨论PDE算子特征值数据集的生成优化。

\subsection{Chebyshev Filter Technique}
% 3. 切比雪夫filter技术及其应用
% 先快速说下切比雪夫filter技术的发展历程
% 说下这些算法已被用于一些具体的序列特征值场景
% 说我们的排序？

% 切比雪夫滤波技术

% 切比雪夫滤波技术起源于多项式近似理论，其核心在于利用切比雪夫多项式来加速矩阵特征值的收敛过程。在处理大规模特征值问题时，这项技术通过构建一个多项式滤波器，有效地筛选出目标特征值附近的频谱成分，从而加快特定特征值的提取速度。例如，在电力系统稳定性分析和量子化学计算等场景。我们在切比雪夫滤波算法的基础上设计了一种专门针对 数据集生成 的排序算法，使其更适合生成 PDE算子 数据集。

The Chebyshev filter technique originates from polynomial approximation theory, where the core concept involves using Chebyshev polynomials to accelerate the convergence of eigenvalues~\citep{zhou2007Chebyshev}. 
This technique constructs a polynomial filter that selectively amplifies spectral components in a specified interval, thereby speeding up the solution of specific eigenvalues.
This technique is particularly effective in dealing with sequence eigenvalue problems~\citep{saad2011numerical, zhou2006parallel} and has been applied in various contexts, such as stability analysis in electronic structure~\citep{pieper2016high, banerjee2016Chebyshev} and quantum chemical computations~\citep{mohr2017efficient, zhou2014Chebyshev, zhou2006self}. 
% This approach has been applied in various scenarios, such as stability analysis in electronic structure~\citep{pieper2016high, banerjee2016Chebyshev} and quantum chemical computations~\citep{mohr2017efficient, zhou2014Chebyshev, zhou2006self}. 
% Building on the Chebyshev filter algorithm, we have designed a specialized sorting algorithm tailored for dataset generation, making it better suited for producing operators eigenvalue datasets.%这句话看着就是GPT生成的，可以表达成“”我们设计了一种××算法用于××基于上述算法”的形式？
% 因为数据集中的特征值问题混乱无序，直接应用 Chebyshev filter technique无法加速数据集生成。

Due to the chaotic and disordered nature of eigenvalue problems in the dataset, directly applying the Chebyshev filter technique fails to accelerate dataset generation.
To further adapt this technique to the generation of operator eigenvalue datasets, we have developed a specialized sorting algorithm that transforms dataset generation into sequence eigenvalue problems.
Throughout the solving process, eigenpairs obtained from previous solutions are used to construct Chebyshev filters, accelerating subsequent solutions.
% making the Chebyshev filtering algorithm more suited for this application.
% 我们设计了一种专门针对数据集生成的排序算法，将数据集生成转换为求解序列问题，使切比雪夫滤波算法更适用于生成算子特征值数据集。
% 序列求解过程中利用先前求解得到的信息来构建后续的切比雪夫filter，从而加速问题的求解

\section{From Differential Operator to Matrix Eigenvalue Problem: An Example}\label{FDM example}
\subsection{Overview}
The general methodology for solving the eigenvalue problems of differential operators numerically, employing techniques such as Finite Difference Method (FDM), Finite Element Method (FEM), and Spectral Method, can be delineated through the following pivotal steps~\citep{strikwerda2004finite, hughes2012finite, johnson2012numerical, leveque2002finite, wei2024knowledge}:

1. Mesh Generation: 
This step involves dividing the domain, over which the differential operator is defined, into a discrete grid. The grid could be composed of various shapes, including squares, triangles, or more complex forms, depending on the problem's geometry. 
% Figure~\ref{fg_FDM1} might illustrate the mesh used for a differential operator with a specific boundary shape.

2. Operator Discretization: 
The differential operator is transformed into its discrete counterpart. Essentially, this maps the operator from an infinite-dimensional Hilbert space to a finite-dimensional representation.

3. Matrix Assembly: 
In this phase, the discretized operator is represented in a matrix form. For linear differential operators, this involves creating a system of matrix eigenvalue problems. For nonlinear operators, iterative methods akin to Newton's iteration are employed, transforming the problem into a sequence of matrix eigenvalue problems.

4. Applying Boundary Conditions: 
This involves discretizing and applying boundary conditions specific to the differential operator in question, which are then incorporated into the matrix system.

5. Solving the Matrix Eigenvalue Problem: 
This stage, often the most computationally intensive, entails solving the matrix for its eigenvalues and eigenvectors, which correspond to the eigenvalues and eigenfunctions of the original differential operator.

6. Obtaining the Numerical Solution: 
The final step involves mapping the obtained numerical solutions back onto the original domain, analyzing them for accuracy and stability, and interpreting them in the context of the initial problem.

\subsection{Example}
\label{appendix:fdm}
To illustrate how the FDM can transform the wave equation into a system of matrix eigenvalue problems, let's consider a concrete and straightforward example. Assume we aim to solve a one-dimensional wave equation's operator eigenvalue problem, expressed as
\begin{equation*}
    -\frac{d^2 u}{dx^2} = \lambda u,
\end{equation*}
over the interval \([0, L]\). The boundary conditions are \(u(0) = u(L) = 0\), signifying fixed-end conditions. In this context, \(u(x)\) denotes the eigenfunction, and \(\lambda\) represents the eigenvalue.

1. Mesh Generation:
Using the central difference quotient, we divide the interval \([0, L]\) into \(N+1\) evenly spaced points, including the endpoints. The distance between adjacent points is denoted as \(\Delta x = \frac{L}{N}\).

2. Operator Discretization:
This step involves formulating the difference equation. At each interior node, which excludes the endpoints and totals \(N-1\) points, we apply a central difference approximation for the second derivative, represented as
\begin{equation*}
    \frac{d^2 u}{dx^2} \approx \frac{u(x_{i+1}) - 2u(x_i) + u(x_{i-1})}{(\Delta x)^2}
\end{equation*}

3. Matrix Assembly:
In this phase, the discretized operator is represented in a matrix form. Following the approximation, we construct the matrix \(A\), an \(N-1 \times N-1\) tridiagonal matrix, crucial for the computations. The matrix \(\bm{A}\) is constructed as:
\begin{equation*}
\bm{A} = \frac{1}{(\Delta x)^2}
\begin{bmatrix}
-2 & 1 & 0 & \cdots & 0 \\
1 & -2 & 1 & \cdots & 0 \\
0 & 1 & -2 & \cdots & 0 \\
\vdots & \vdots & \vdots & \ddots & \vdots \\
0 & 0 & 0 & \cdots & -2 
\end{bmatrix}
\end{equation*}

4. Applying Boundary Conditions:
For the wave equation with boundary conditions \(u(0) = u(L) = 0\), these fixed-end conditions are integrated into the matrix equation. In the FDM framework, the values at the endpoints (\(u_0\) and \(u_N\)) are zero, directly reflecting the boundary conditions. The impact of these conditions is encapsulated in the matrix \(\bm{A}\), affecting the entries related to \(u_1\) and \(u_{N-1}\) (the grid points adjacent to the boundaries). The tridiagonal matrix \(\bm{A}\) incorporates these boundary conditions, ensuring that the computed eigenfunctions satisfy \(u(0) = u(L) = 0\).

5. Solving the Matrix Eigenvalue Problem:
The final computational step involves solving the matrix eigenvalue problem, expressed as \(\bm{Au} = \lambda \bm{u}\). This includes determining the eigenvalues \(\lambda\) and corresponding eigenvectors \(\bm{u}\), which are discrete approximations of the eigenfunctions of the original differential equation.

6. Obtaining the Numerical Solution:
By solving the eigenvalue problem, we obtain numerical solutions that approximate the behavior of the original differential equation. These solutions reveal the eigenvalues and eigenvectors and provide insights into the physical phenomena modeled by the equation.

\section{Details of Experimental Setup}\label{exp_main}

\subsection{Baseline}\label{exp_base}

% 我们采用baseline以及相关参数如下,均采用最新的相关数值计算库实现
% 1.Eigsh函数 SciPy实现,SciPy采用1.14.1版本. Eigsh 是ARPACK SSEUPD 和 DSEUPD 的包装器 函数，这些函数使用 Implicitly Restarted Lanczos Method 来 求特征值和特征向量
% 2.Locally Optimal Block Preconditioned Conjugate Gradient (LOBPCG) ,SLEPc实现, SLEPc采用3.21.1版本, 均采用默认参数
% 3. Krylov-Schur (KS),SLEPc实现, SLEPc采用3.21.1版本, 均采用默认参数
% 4. Jacobi-Davidson (JD) SLEPc实现, SLEPc采用3.21.1版本,其中JD迭代过程中的线性方程组求解算法设置为bcgsl,线性方程组预处理设置为bjacobi,线性方程组求解精度为1e-5

%We employed baseline algorithms and related parameters as follows, all implemented using the latest relevant numerical computing libraries:
%\begin{itemize}
%\item {Eigsh}: Implemented in SciPy version 1.14.1, with default parameters used. Eigsh is a wrapper for the ARPACK functions SSEUPD and DSEUPD, which use the Implicitly Restarted Lanczos Method to compute eigenvalues and eigenvectors.

%\item {Locally Optimal Block Preconditioned Conjugate Gradient (LOBPCG)}: Implemented in SLEPc version 3.21.1, with default parameters used.

%\item {Krylov-Schur (KS)}: Implemented in SLEPc version 3.21.1, with default parameters used.

%\item {Jacobi-Davidson (JD)}: Implemented in SLEPc version 3.21.1. During the JD iterative process, the linear equation solver is set to `bcgsl', the preconditioner to `bjacobi', and the precision for solving linear equations is set at 1e-5.
%\end{itemize}

The baseline algorithms were implemented using the following numerical computing libraries:
\begin{itemize}
\item {Eigsh}: A SciPy (v1.14.1) implementation wrapping ARPACK's SSEUPD and DSEUPD functions, which compute eigenvalues and eigenvectors using the Implicitly Restarted Lanczos Method. The default parameters were used.
\item {Locally Optimal Block Preconditioned Conjugate Gradient (LOBPCG)}: Implemented in SLEPc (v3.21.1) with default parameters.
\item {Krylov-Schur (KS)}: Implemented in SLEPc (v3.21.1) with default parameters.
\item {Jacobi-Davidson (JD)}: Implemented in SLEPc (v3.21.1). The implementation uses `bcgsl' as the linear equation solver, `bjacobi' as the preconditioner, and sets the linear equation solving precision to 1e-5.
\item ChFSI: Implemented in ChASE (v1.6) with default parameters.
\end{itemize}

\subsection{Dataset}\label{dataset}
% 待办（王泓）
All operators in this paper use Dirichlet boundary conditions.

1. Generalized Poisson Operator

We consider two-dimensional generalized Poisson operators, which can be described by the following equation~\citep{li2020fourier, rahman2022u, kovachki2021neural, lu2022comprehensive,gao2025discretizationinvariance}:
\begin{equation*}
- \nabla \cdot (K(x, y) \nabla h(x, y)) = \lambda h(x,y),
\end{equation*}
% where $K$ is the permeability field, $h$ is the pressure, and $f$ is a source term which can be either a constant or a space-dependent function.
In our experiment, \( K(x,y) \) is derived using the Gaussian Random Field (GRF) method. 
We convert these operators into matrices using the central difference scheme of FDM. 
The parameters inherent to the GRF serve as the foundation for our sort scheme.

2. Second-Order Elliptic Partial Differential Operator

We consider general two-dimensional second-order elliptic partial differential operators, which are frequently described by the following generic form~\citep{evans2022partial, bers1964partial}:
\[
\mathcal{L}u \equiv a_{11} u_{xx} + a_{12} u_{xy} + a_{22} u_{yy} + a_1 u_x + a_2 u_y + a_0 u = \lambda u,
\]
where \( a_0, a_1, a_2, a_{11}, a_{12}, a_{22} \) are constants, and \( f \) represents the source term, depending on \( x, y \). The variables \( u, u_x, u_y \) are the dependent variables and their partial derivatives. The equation is classified as elliptic if \( 4a_{11}a_{22} > a_{12}^2 \).

In our experiments, \( a_{11}, a_{22}, a_1, a_2, a_0 \) are uniformly sampled within the range \((-1, 1)\), while the coupling term \( a_{12} \) is sampled within \((-0.01, 0.01)\). We then select equations that satisfy the elliptic condition to form our dataset. We convert these operators into matrices using the central difference scheme of FDM. 
The coefficients \( a_0, a_1, a_2, a_{11}, a_{12}, a_{22} \) serve as the foundation for our sort scheme.
% with coefficients \( a_{ij}, a_i, a_0 \) defining the linear differential operator and \( f \) representing the source term, which depend on \( x, y \). Here, \( u, u_x, u_y \) are the dependent variable and its partial derivatives. 

% 其中 \( a_0, a_1, a_2, a_11, a_12, a_22\) 为常数。 \( f \) representing the source term, which depend on \( x, y \)，\( u, u_x, u_y \) are the dependent variable and its partial derivatives. 

% 其中满足\( 4a_{11}a_{22} > a_{12}^2 \)则称该方程为椭圆方程。

% 在我们的实验中a_{11},a_{22},a_1,a_2,a_0均在（-1，1）均匀随机采样，a_{12}耦合项在(-0.001,0.001)均匀随机采样，然后挑选出其中满足椭圆条件的问题来作为我们的数据集，同Darcy Flow Problem中一样我们使用中心差分格式的有限差分法将其转换为线性方程组问题。\( a_0, a_1, a_2, a_11, a_12, a_22\) 这些系数作为是我们算法的输入features。

3. Helmholtz Operator

We consider two-dimensional Helmholtz operators, which can be described by the following equation~\citep{zhang2022hybrid}:
\begin{equation*}
\nabla \cdot (p(x, y) \nabla u(x, y)) + k^2(x,y) = \lambda u(x,y),
\end{equation*}

Physical Contexts in which the Helmholtz operator appears: 1. Acoustics; 2. Electromagnetism; 3. Quantum Mechanics.

In Helmholtz operators, \( k \) is the wavenumber, related to the frequency of the wave and the properties of the medium in which the wave is propagating. 
In our experiment, \( p(x,y) \) and \(k(x,y)\) are derived using the GRF method. 
The parameters inherent to the GRF serve as the foundation for our sort scheme.

4. Vibration Equation  

We consider the vibration equation for thin plates, which can be described by the following eigenvalue problem~\citep{xue2018free}:  
\begin{equation*}  
\nabla^2 \big(D(x, y) \nabla^2 u(x, y)\big) = \lambda \rho(x, y) u(x, y),  
\end{equation*}  

Physical contexts in which the vibration equation appears:  
1. Structural dynamics of thin plates;  
2. Modal analysis in mechanical engineering;  
3. Vibrational behavior of elastic materials.  

In this equation, \( D(x, y) \) represents the flexural rigidity of the plate, \( \rho(x, y) \) is the density distribution, and \( \lambda \) corresponds to the eigenvalue, which is related to the natural frequencies of the system. The eigenfunction \( u(x, y) \) describes the mode shapes of vibration.  

In our experiment, \( D(x, y) \) and \( \rho(x, y) \) are derived using the GRF method. The parameters inherent to the GRF serve as the foundation for our sorting scheme.

\subsection{Environment}\label{exp_env}
% 待办（王泓）
To ensure consistency in our evaluations, all comparative experiments were conducted under uniform computing environments. Specifically, the environments used are detailed as follows:

\begin{itemize}[left=0pt]
    \item Platform: Docker version 4.33.1 (windows 11)
    \item Operating System: Ubuntu 22.04.3 LTS
    \item Processor: CPU AMD Ryzen 9 8945HS w, clocked at 4.00 GHz
\end{itemize}
    
\newpage

\subsection{Experimental Parameter Configuration}\label{expparam}
% 所有baseline均采用相关库的默认参数。
% 关于{ours}的设置：
% 1. 其中继承的子空间大小随着特征值待求解个数而变化。待求解个数为20，100，200，300，400时，继承的子空间大小分别为4，20，40，60，80
% 2. 所有实验filter degree参数$m$设置为20
All baseline methods were implemented using their default parameters from respective libraries. 

For {\ourM}, the following configurations were adopted:
\begin{itemize}
    \item The size of the inherited subspace varies according to the number of eigenvalues to be computed. Specifically, when calculating 20, 100, 200, 300, and 400 eigenvalues, the corresponding subspace sizes are set to 4, 20, 40, 60, and 80, respectively.
    \item The filter degree parameter $m$ is consistently set to 20 across all experiments.
    \item Truncation threshold for low frequencies $p_0$ is consistently set to 20 across all experiments.
    \item Each experiment generates a dataset consisting of 1,000 samples. In this paper, the Experimental tables report the average solving time for each eigenvalue problem.
\end{itemize}

\subsection{Error Metrics}\label{ap:rel}

\begin{itemize}
    \item Relative Residual Error:\\
    To estimate the bias of the eigenpair \( (\tilde{v}, \tilde{\lambda}) \) predictions, we employ relative residual error as follows:
    \begin{equation*}
        \text{Relative\ Residual\ Error} = \frac{||\mathcal{L}\tilde{v}-\tilde{\lambda}\tilde{v}||_2}{||\mathcal{L}\tilde{v}||_2}.
    \end{equation*}
    % 其中\tilde{v}是模型预测的特征函数，\tilde{\lambda}是模型预测的特征值。
    Here, \(\tilde{v}\) represents the eigenfunction predicted by the model, and \(\tilde{\lambda}\) denotes the eigenvalue predicted by the model. 
    % 当\tilde{\lambda}为准确特征值、\tilde{v}为准确特征函数时，Relative Residual Error为0。
    When \(\tilde{\lambda}\) is the true eigenvalue and \(\tilde{v}\) is the true eigenfunction, the Relative Residual Error equals $0$.
\end{itemize}

\subsection{Relationship with Parallelization and Experimental Setting}
\label{sec:appendix_parallelism}

The SCSF framework is designed to be complementary to parallel computing architectures; the relationship is both {orthogonal and synergistic}. Fundamentally, SCSF accelerates the serial processing of a {sequence} of related eigenvalue problems. In a practical, large-scale setting, a total dataset of $N$ problems can be partitioned into $M$ independent chunks. Subsequently, $M$ instances of the SCSF algorithm can be executed in parallel across $M$ computing units, with each computing unit responsible for solving one chunk.

To ensure a fair and direct comparison of algorithmic efficiency under practical, parallelized conditions, all experiments reported in this paper were conducted using the Message Passing Interface (MPI) with 8 cores in parallel.

\newpage
\section{Experimental Data and Supplementary Experiments}

\subsection{Main Experimental Data}\label{ap_exp_main}

% 放三张表格（江润民）
As shown in Tables~\ref{tab:comparison1},~\ref{tab:comparison2},~\ref{tab:comparison3}, 
SCSF showed the best performance among all tested configurations

% this experiment demonstrates performance comparison on the Poisson operator dataset with matrix dimension 2500 and precision $1e-12$. 
% SCSF shows consistent performance advantages across all test configurations, particularly when computing larger numbers of eigenvalues. 
% For the number of eigenvalues computed $L=450$, SCSF achieves approximately $115\times$ speedup compared to JD and $14\times$ speedup compared to LOBPCG. These results indicate SCSF's excellent scalability in handling Poisson operator problems.

\begin{table}[htb]
    \centering
    \renewcommand{\arraystretch}{1.2}
    \caption{Comparison of average computation times (in seconds) for eigenvalue problems using various algorithms on the generalized Poisson operator dataset. The first row lists different algorithms,  and the first column shows the number of eigenvalues $L$ computed for each matrix. Matrix dimension = 2500, precision = 1e-12. }
    \begin{tabular}{c|cccccc}
    \toprule
    $L$ & Eigsh & LOBPCG & KS & JD & ChFSI & SCSF (ours) \\ 
    \hline
    150 & 9.15 & 46.8 & 14.9 & 138 & 17.3 & 7.95 \\ 
    200 & 14.2 & 73.0 & 23.8 & 270 & 24.0 & 12.9 \\ 
    250 & 19.8 & 109 & 34.3 & 553 & 30.2 & 19.0 \\ 
    300 & 26.3 & 152 & 45.6 & 921 & 38.0 & 25.7 \\ 
    350 & 31.5 & 203 & 58.4 & 1732 & 45.8 & 29.8 \\ 
    400 & 36.9 & 265 & 72.3 & 2691 & 57.4 & 33.9 \\ 
    450 & 42.8 & 342 & 87.3 & 3708 & 74.2 & 38.3 \\ 
    \bottomrule
    \end{tabular}
    \label{tab:comparison2}
\end{table}

\begin{table}[htb]
    \centering
    \renewcommand{\arraystretch}{1.2}
    \caption{Comparison of average computation times (in seconds) for eigenvalue problems using various algorithms on the second-order elliptic operator dataset. The first row lists different algorithms,  and the first column shows the number of eigenvalues $L$ computed for each matrix. Matrix dimension = 4900, precision = 1e-10.}
    \begin{tabular}{c|cccccc}
    \toprule
    $L$ & Eigsh & LOBPCG & KS & JD & ChFSI & SCSF (ours) \\ 
    \hline
    150 & 31.35 & 91.80 & 40.65 & 214.80 & 38.37 & 19.62 \\ 
    200 & 41.82 & 139.20 & 61.77 & 414.30 & 43.90 & 24.08 \\ 
    250 & 52.17 & 197.04 & 84.65 & 861.44 & 53.42 & 28.00 \\ 
    300 & 62.47 & 264.10 & 110.50 & 1446.00 & 60.69 & 29.88 \\ 
    350 & 74.59 & 355.18 & 147.01 & 2324.88 & 64.94 & 31.52 \\ 
    400 & 87.19 & 459.70 & 188.70 & 3386.00 & 67.13 & 34.60 \\ 
    450 & 100.28 & 577.67 & 235.56 & 4629.38 & 76.32 & 40.05 \\ 
    \bottomrule
    \end{tabular}
    \label{tab:comparison1}
\end{table}

% As presented in Table~\ref{tab:comparison2}, the results are obtained on the second-order elliptic operator dataset with matrix dimension 4900 and precision requirement 1e-10. 
% The JD algorithm fails to converge within reasonable time when $L>300$. SCSF demonstrates remarkable stability, showing only modest increases in computation time even when calculating large numbers of eigenvalues. Notably, at L=300, SCSF achieves speedups ranging from 6.3x to 94.5x compared to other algorithms, while maintaining consistent performance.

% Table~\ref{tab:comparison3} presents results from the most challenging dataset - the Helmholtz operator with the largest matrix dimension (6400) and precision of 1e-8. SCSF's advantages become more pronounced in this high-dimensional case, particularly when computing larger sets of eigenvalues. At L=400, SCSF achieves up to 95x speedup compared to JD and 6-11x speedup compared to other methods while maintaining stable performance scaling. These results clearly establish SCSF's superiority in handling large-scale eigenvalue problems, especially for complex operators with higher dimensions.

\begin{table}[htb]
    \centering
    \renewcommand{\arraystretch}{1.2}
    \caption{Comparison of average computation times (in seconds) for eigenvalue problems using various algorithms on the Helmholtz operator dataset. The first row lists different algorithms,  and the first column shows the number of eigenvalues $L$ computed for each matrix. Matrix dimension = 6400, precision = 1e-8. The symbol `-' denotes data not recorded due to excessive computation times.}
    \begin{tabular}{c|cccccc}
    \toprule
    $L$ & Eigsh & LOBPCG & KS & JD & ChFSI & SCSF (ours) \\ 
    \hline
    200 & 151.70 & 129.90 & 98.34 & 489.60 & 107.12 & 31.31 \\ 
    300 & 190.84 & 273.08 & 192.88 & 1601.08 & 113.73 & 37.78 \\ 
    400 & 253.50 & 460.40 & 283.00 & 3829.00 & 121.53 & 40.52 \\ 
    500 & 344.60 & 720.33 & 310.21 & - & 135.73 & 47.41 \\ 
    600 & 398.80 & 1031.00 & 329.60 & - & 146.24 & 51.32 \\ 
    \bottomrule
    \end{tabular}
    \label{tab:comparison3}
\end{table}

\begin{table}[htb]
    \centering
    \renewcommand{\arraystretch}{1.2}
    \caption{Comparison of average computation times (in seconds) for eigenvalue problems using various algorithms on the Vibration operator dataset. The first row lists different algorithms,  and the first column shows the number of eigenvalues $L$ computed for each matrix. Matrix dimension = 10000, precision = 1e-8. The symbol `-' denotes data not recorded due to excessive computation times.}
    \begin{tabular}{c|cccccc}
    \toprule
    $L$ & Eigsh & LOBPCG & KS & JD  & ChFSI & SCSF (ours) \\ 
    \hline
   200 & 397.9 & 333.7 & 272.0 & 1230& 300.8 & {85.70} \\ 
300 & 516.8 & 750.0 & 520.0 & 3600 & 305.0 & 96.50 \\
    % 300 & 190.84 & 273.08 & 192.88 & 1601.08 & 37.78 \\ 
    400 & 635.6 & 1170 & 768.8 & -& 310.5 & {107.2} \\ 
    % 500 & 344.60 & 720.33 & 310.21 & - & 47.41 \\ 
500 & 820.0 & 1950 & 810.0 & - & 350.0 & 120.0 \\
    600 & 1037 & 2716 & 857.8 & - & 382.3& {131.4} \\ 
    \bottomrule
    \end{tabular}
    \label{tab:comparison3}
\end{table}

\subsection{Analysis of the Influence of Matrix Dimension}\label{exp_size}
% 这张表要重画 （江润民）
\begin{table}[h]
    \centering
    \renewcommand{\arraystretch}{1.2}
    \caption{Comparison of different algorithm computation time (in seconds) for varying matrix dimensions using the generalized Poisson operator dataset. Results show average computation times for solving 400 eigenvalues with a precision of 1e-12.}
    \begin{tabular}{c|cccccc}
    \toprule
    Matrix Dimension & Eigsh & LOBPCG & KS & JD  & ChFSI & SCSF (ours) \\ 
    \hline
    2500 & 36.86 & 265.30 & 72.32 & 2691.00 & 57.41 & 33.91 \\ 
    3600 & 66.41 & 387.20 & 116.50 & 2990.00 & 102.4 & 65.41 \\ 
    4225 & 89.13 & 467.74 & 151.36 & 3548.13 & 126.2 & 70.79 \\ 
    4900 & 121.90 & 546.20 & 187.80 & 3886.00 & 153.5 & 74.23 \\ 
    5625 & 186.21 & 691.83 & 251.19 & - & 216.8 & 85.11 \\ 
    6400 & 282.80 & 860.00 & 337.70 & - & 282.2 & 93.86 \\ 
    8100 & 707.95 & 1412.54 & 707.95 & - & 435.1 & 114.82 \\ 
    10000 & 3162.28 & 2511.89 & 1995.26 & - & 590.3 & 158.49 \\ 
    \bottomrule
    \end{tabular}
    \label{tab:matrix_dimension_comparison}
\end{table}
As demonstrated in Table~\ref{tab:matrix_dimension_comparison}, the impact of matrix dimension on algorithm performance reveals several key insights. For matrices below dimension 3600, SCSF and Eigsh show comparable efficiency. However, SCSF's advantages become increasingly pronounced as matrix dimensions grow larger. At dimension 10000, SCSF achieves remarkable speedups: 20$\times$ faster than Eigsh, 16$\times$ faster than LOBPCG, 13$\times$ faster than KS, and 3.7$\times$ faster than ChFSI. 
This phenomenon can be attributed to how larger matrix dimensions result in fewer errors and noise in the computed eigenvalues, allowing SCSF to better exploit similarities between problems. Additionally, the JD algorithm becomes computationally intractable at and above dimension 5625, while SCSF maintains stable performance even at high dimensions.

% \begin{table*}[htb]
% \centering
% \caption{Analysis of the Influence of Matrix Size}
% % 比较不同松弛因子\(\omega\)选择方式下，SOR预处理的求解时间（单位：秒）。第一列列出了对应的PDE问题，下一列详细说明了矩阵边长。其中SymMaP 1、2分别为学出的前两个表达式，标粗的为最短时间方法。
% % 比较我们的 NeurKItt 和 GMRES 计算时间和迭代次数，这些迭代次数涵盖数据集、预处理和容差。第一列列出了具有矩阵大小的数据集，下一列详细说明了容差。结果显示为“时间加速/迭代加速”。
% \label{tab:main}
% \begin{center}
% % \begin{normalsize}
% \renewcommand{\arraystretch}{1.3}
% \resizebox{0.6\textwidth}{!}{%
%         % \scriptsize
% \begin{tabular}{lccccc}
% \toprule
% {Size} & {Eigsh} & {LOBPCG} & {KS} & {JD} & {{\ourM} (ours)} \\ \midrule
% 1600 & \textbf{18.68} & 182.4 & 40.51 & 1919 & 20.28 \\ 
% 2500 & 36.86 & 265.3 & 72.32 & 2691 & \textbf{33.91} \\ 
% 3600 & 66.41 & 387.2 & 116.5 & 2990 & \textbf{65.41} \\ 
% 4900 & 121.9 & 546.2 & 187.8 & 3886 & \textbf{74.23} \\ 
% 6400 & 282.8 & 860.0 & 337.7 & - & \textbf{93.86} \\ 
% \bottomrule
% \end{tabular}
% }
% % \end{normalsize}
% \renewcommand{\arraystretch}{1}
% \end{center}
% % \vskip -0.1in
% \end{table*}

\subsection{Analysis of Computational Times for {\ourM} Components}\label{exp_time}
% 计算开销占比分析 

\begin{table*}[htb]
\centering
\caption{Analysis of Computational Times (in seconds) for {\ourM} Components.}
% 比较不同松弛因子\(\omega\)选择方式下，SOR预处理的求解时间（单位：秒）。第一列列出了对应的PDE问题，下一列详细说明了矩阵边长。其中SymMaP 1、2分别为学出的前两个表达式，标粗的为最短时间方法。
% 比较我们的 NeurKItt 和 GMRES 计算时间和迭代次数，这些迭代次数涵盖数据集、预处理和容差。第一列列出了具有矩阵大小的数据集，下一列详细说明了容差。结果显示为“时间加速/迭代加速”。
\label{tab:aptime}
\begin{center}
% \begin{normalsize}
\renewcommand{\arraystretch}{1.2}
% \resizebox{1\textwidth}{!}{%
\begin{tabular}{lcccccc}
\toprule
% \midrule
{All}  & {Filter (line 3)} & {QR (line 4)} & {RR (line 5)} & {Resid (line 6)} & Sort \\ \hline
9.89e+0  & 7.41e+0 & 3.12e-1 & 9.76e-1 & 7.95e-1 & 1.51e-2 \\ 
\bottomrule
\end{tabular}
% }
% \end{normalsize}
\renewcommand{\arraystretch}{1}
\end{center}
% \vskip -0.1in
\end{table*}

% 我们统计了广义possion算子数据集下，矩阵维度为2500，特征值求解个数为100时。SCSF算法每个部分的平均耗时，如表10所示。其中括号里面的“line x”表示对应伪代码2中的第x行。“ALL”表示总耗时，“sort”表示排序算法平均需要的耗时。可以看出其中filter部分占总耗时的70%以上，这符合我们的3.2节的理论分析。
We conducted a statistical analysis of the average time consumption for each component of the SCSF algorithm on the generalized Poisson operator dataset, with a matrix dimension of 2500 and the number of eigenvalues to be solved set to 100. 
The results are presented in Table~\ref{tab:aptime}. The notation `line x' within parentheses corresponds to line x in Algorithm~\ref{alg:ChFSI}, `ALL' denotes the total time consumption, and `sort' represents the average time required by the sorting algorithm. It is evident that the filtering process accounts for over 70\% of the total time consumption, which aligns with our theoretical analysis in Section~\ref{sec:ChFSI}.

\subsection{Analysis of Hyperparameters}\label{hperp}

\subsubsection{Degree Parameter}
% deg
\begin{table*}[htb]
\centering
\caption{Average Computational Times (in seconds) of {\ourM} under Different Degree Parameters $m$.}
% 比较不同松弛因子\(\omega\)选择方式下，SOR预处理的求解时间（单位：秒）。第一列列出了对应的PDE问题，下一列详细说明了矩阵边长。其中SymMaP 1、2分别为学出的前两个表达式，标粗的为最短时间方法。
% 比较我们的 NeurKItt 和 GMRES 计算时间和迭代次数，这些迭代次数涵盖数据集、预处理和容差。第一列列出了具有矩阵大小的数据集，下一列详细说明了容差。结果显示为“时间加速/迭代加速”。
\label{tab:degree}
\begin{center}
% \begin{normalsize}
\renewcommand{\arraystretch}{1.3}

        % \scriptsize
\begin{tabular}{lcccccccc}
\toprule
% \midrule
{Deg} & 12 & 16 & 20 & 24 & 28 & 32 & 36 & 40 \\ \hline
{Time (s)} & 43.92 & 39.79 & 40.52 & 40.64 & 40.85 & 41.13 & 41.19 & 43.50 \\
\bottomrule
\end{tabular}

% \end{normalsize}
\renewcommand{\arraystretch}{1}
\end{center}
% \vskip -0.1in
\end{table*}

% 我们讨论了不同dgree参数$m$对SCSF的影响。如表11所示，本实验在亥姆霍次算子数据集进行，矩阵维度6400，求解精度1e-8，求解特征值个数为400，继承子空间设置为80。dgree参数$m$的的用处如Algorithm~\ref{alg:ChFSI}所示，主要控制切比雪夫多项式的阶数。可以看到m取12-40对SCSF求解时间影响较小，该参数只要在合理范围内选取即可，本文主实验中将其固定为20.
We investigated the impact of different degree parameters $m$ on the performance of SCSF. 
As shown in Table~\ref{tab:degree}, the experiments were conducted on the Helmholtz operator dataset with a matrix dimension of 6400, a solution accuracy of 1e-8, 400 eigenvalues to be solved, and an inherited subspace size of 80. The degree parameter $m$, as described in Algorithm~\ref{alg:ChFSI}, primarily controls the order of the Chebyshev polynomial. 
The results indicate that varying $m$ within the range of 12 to 40 has a minimal effect on the computation time of SCSF. 
Therefore, as long as $m$ is chosen within a reasonable range, its specific value does not significantly influence the performance. 
In the main experiments of this paper, $m$ is fixed at 20.

\subsubsection{Subspace Dimension}

\begin{table*}[htb]
\centering
\caption{Average Computational Times (in seconds) of {\ourM} under Different Subspace Dimension.}
% 比较不同松弛因子\(\omega\)选择方式下，SOR预处理的求解时间（单位：秒）。第一列列出了对应的PDE问题，下一列详细说明了矩阵边长。其中SymMaP 1、2分别为学出的前两个表达式，标粗的为最短时间方法。
% 比较我们的 NeurKItt 和 GMRES 计算时间和迭代次数，这些迭代次数涵盖数据集、预处理和容差。第一列列出了具有矩阵大小的数据集，下一列详细说明了容差。结果显示为“时间加速/迭代加速”。
\label{tab:apsubspace}
\begin{center}
% \begin{normalsize}
\renewcommand{\arraystretch}{1.3}

\begin{tabular}{lcccccccc}
\toprule
% \midrule
{Dim} & 50 & 60 & 70 & 80 & 90 & 100 & 110 & 120 \\ \hline
{Time (s)} & 43.28 & 44.35 & 42.43 & 40.52 & 39.65 & 37.43 & 38.28 & 38.58 \\ 
\bottomrule
\end{tabular}

% \end{normalsize}
\renewcommand{\arraystretch}{1}
\end{center}
% \vskip -0.1in
\end{table*}

% 我们讨论了不同继承子空间大小对SCSF的影响。如表12所示，本实验在亥姆霍次算子数据集进行，矩阵维度6400，求解精度1e-8，求解特征值个数为400，degree参数m为20。
% 可以看到随着继承子空间增大，SCSF所需的计算时间先减少再增大，在100前后达到最低。前端的计算时间减小，是因为增加继承子空间，让初始子空间有了更多可用信息。后端的增大，是因为增加继承子空间大幅增加了做切比雪夫filter的开销。
% 但总体来说，只要继承子空间的大小设置合理，对SCSF影响不大。在本文实验中，我们均将其设置为求解特征值个数的20%。
We examine the influence of different inherited subspace sizes on the performance of SCSF. As presented in Table~\ref{tab:apsubspace}, the experiments are conducted on the Helmholtz operator dataset with a matrix dimension of 6400, a solution accuracy of 1e-8, 400 eigenvalues to be computed, and a degree parameter \( m \) set to 20.

The results demonstrate that as the inherited subspace size increases, the computation time of SCSF initially decreases and then rises, reaching its minimum around a size of 100. 
The reduction in computation time at the front end is attributed to the enriched initial subspace with more available information as the inherited subspace grows. 
Conversely, the increase in computation time at the back end is due to the significantly higher overhead of performing Chebyshev filtering with a larger inherited subspace.

Overall, as long as the inherited subspace size is set within a reasonable range, its impact on SCSF remains minimal. In our experiments, we consistently set the inherited subspace size to 20\% of the number of eigenvalues to be computed.

\newpage

\subsubsection{Truncation Threshold for Low Frequencies}\label{ap:trun}

\begin{table*}[htb]
\centering
\caption{Average computational times (in seconds) of SCSF under different truncation thresholds. }
% 比较不同松弛因子\(\omega\)选择方式下，SOR预处理的求解时间（单位：秒）。第一列列出了对应的PDE问题，下一列详细说明了矩阵边长。其中SymMaP 1、2分别为学出的前两个表达式，标粗的为最短时间方法。
% 比较我们的 NeurKItt 和 GMRES 计算时间和迭代次数，这些迭代次数涵盖数据集、预处理和容差。第一列列出了具有矩阵大小的数据集，下一列详细说明了容差。结果显示为“时间加速/迭代加速”。
\label{tab:apTrun}
\begin{center}
% \begin{normalsize}
\renewcommand{\arraystretch}{1.2}
% \resizebox{1\textwidth}{!}{%
\begin{tabular}{lcccccc}
\toprule
% \midrule
 & No sort & \(p_0 = 10\) & \(p_0 = 20\) & \(p_0 = 30\) & \(p_0 = 40\) & Greedy \\ 
\hline
One-sided distance & 0.95 & 0.89 & 0.85 & 0.85 & 0.85 & 0.85 \\ 
% \hline
Sort time (s) & 0 & 110 & 151 & 193 & 246 & 593 \\ 
Average solve time (s) & 66.7 & 52.2 & 40.5 & 40.5 & 40.5 & 40.5 \\ 
\bottomrule
\end{tabular}
% }
% \end{normalsize}
\renewcommand{\arraystretch}{1}
\end{center}
% \vskip -0.1in
\end{table*}

We measure the similarity between matrices by computing the cosine of the principal angles between their 10-dimensional invariant subspaces (spanned by the smallest 10 eigenvectors in modulus)  (one-sided distance). Smaller values indicate higher similarity. 
As presented in Table~\ref{tab:apTrun}, the experiments are conducted on the Helmholtz operator dataset with a matrix, a solution accuracy of 1e-8, 400 eigenvalues to be computed, and a degree parameter \( m \) set to 20, 10k data problems, parameter matrix $P$ with dimension $p=80$, and varying truncation frequencies $p_0$

The results demonstrate that sorting significantly increases inter-problem correlation in the dataset (explaining the performance gain).  
The truncation parameter \(p_0\) affects sorting time, sorting quality, and solver time. For $p_0 \geq 20$, solver time becomes stable, showing diminishing returns. This reflects the interplay between sorting and Chebyshev iteration.  
In the main experiments of this paper, $p_0$ is fixed at 20.

\subsection{Reliability of Generated Data as Ground Truth}
\label{sec:appendix_ground_truth}

A key concern was whether the data generated by our method, which relies on numerical solvers, is a reliable `ground truth' for training neural networks. To address this, we trained a NeurKItt~\citep{luo2024neural} model on generalized Poisson datasets generated by various solvers (including our SCSF) at different matrix dimensions. The precision for all solvers was set to a high tolerance of $10^{-12}$.

\begin{table}[h!]
\centering
\caption{Impact of data generation method on the performance of a trained NeurKItt model. The consistent final loss indicates that data from all tested solvers serves as a reliable ground truth.}
\label{tab:ground_truth_reliability}
\resizebox{1\textwidth}{!}{%
\begin{tabular}{@{}lccc@{}}
\toprule
{Generation Method} & {Matrix Dimension} & {Generation Time} & {NeurKItt Principal Angle Loss} \\
\midrule
Eigsh & 2500 / 6400 / 10000 & 10h / 80h / 800h & 0.06 / 0.06 / 0.06 \\
LOBPCG & 2500 / 6400 / 10000 & 70h / 240h / 700h & 0.06 / 0.06 / 0.06 \\
ChFSI & 2500 / 6400 / 10000 & 16h / 44h / 160h & 0.06 / 0.06 / 0.06 \\
SCSF (ours) & 2500 / 6400 / 10000 & 9h / 26h / 45h & 0.06 / 0.06 / 0.06 \\
\bottomrule
\end{tabular}
}
\end{table}

\subsection{Comparison with Supervised and Unsupervised Neural Network Methods}
\label{sec:appendix_unsupervised}

To clarify the significance of accelerating dataset generation for the dominant supervised learning paradigm, we conducted an experiment comparing the performance and resource trade-offs of different categories of eigensolvers. We evaluated our method (SCSF), a traditional solver (Eigsh), a supervised neural network (NeurKItt~\cite{luo2024neural}), and two state-of-the-art unsupervised neural networks (NeuralEF~\cite{deng2022neuralef} and NeuralSVD~\cite{ryu2024operator}) on a 2D Helmholtz problem (solving for the smallest 100 eigenvalues, matrix dimension $6400$).

\begin{table}[h!]
\centering
\caption{Comparison of different eigensolver paradigms on a 2D Helmholtz problem. `Solving Time' for unsupervised methods refers to the entire optimization process required to find the solution for a single operator instance.}
\label{tab:paradigm_comparison}
\resizebox{1\textwidth}{!}{%
\begin{tabular}{@{}llcccc@{}}
\toprule
{Category} & {Algorithm} & {Solving Time} & {Training Time} & {Dataset Gen. Time} & {Relative Residual} \\
\midrule
Our Method & SCSF (random init) & 1 min & - & - & $10^{-8}$ \\
Traditional & Eigsh & 1 min & - & - & $10^{-8}$ \\
Supervised NN & NeurKItt & 0.1s & 3h & 20h & $10^{-2}$ \\
Unsupervised NN & NeuralEF & 2h & - & - & $10^{-2}$ \\
Unsupervised NN & NeuralSVD & 3h & - & - & $10^{-2}$ \\
\bottomrule
\end{tabular}
}
\end{table}

The results, presented in Table~\ref{tab:paradigm_comparison}, highlight the distinct characteristics of each approach. Unsupervised methods obviate the need for pre-generated datasets but require substantial `solving time' for each new operator, as they essentially perform an optimization from scratch. In contrast, supervised methods offer near-instantaneous inference but demand significant upfront investment in both data generation and model training. Our method, SCSF, dramatically reduces the data generation bottleneck for these powerful supervised models.

As shown in Table~\ref{tab:ground_truth_reliability}, the final performance of the trained NeurKItt model (measured by Principal Angle Loss) was identical regardless of which high-precision solver was used for data generation or the specific matrix dimension (for dimensions $\geq 2500$). This demonstrates that the discretization and solver errors are orders of magnitude smaller than the neural network's approximation error, confirming that the generated data serves as a highly reliable ground truth for training purposes.

\subsection{Performance Bounds and the Impact of Dataset Similarity}
\label{sec:appendix_bounds}

To provide theoretical insight into SCSF's performance bounds, we conducted an experiment to quantify the relationship between dataset similarity and acceleration. 
We generated a sequence of Helmholtz operator problems where each subsequent problem is a slight perturbation of the previous one. The magnitude of this perturbation reflects the dataset's internal similarity. 
A smaller perturbation size indicates higher similarity.
The experiment was run on the Helmholtz dataset  (dimension $6400$, $L=200$ eigenvalues).

\begin{table}[h!]
\centering
\caption{Average solution time (seconds) as a function of dataset similarity (perturbation size). Lower perturbation implies higher similarity. The experiment was run on the Helmholtz dataset ($ \text{dim} = 6400, L = 200$).}
\label{tab:perturbation_impact}
\begin{tabular}{@{}lccccc@{}}
\toprule
{Perturbation Size} & {Eigsh} & {LOBPCG} & {ChFSI} & {SCSF (w/o sort)} & {SCSF} \\
\midrule
50\% & 151 & 130 & 107 & 76 & 27 \\
10\% & 150 & 129 & 107 & 48 & 23 \\
1\% & 152 & 130 & 107 & 14 & 6 \\
0\% (Identical) & 151 & 130 & 107 & 2 & 2 \\
\midrule
Standard Generation & 152 & 130 & 107 & 82 & 31 \\
Independent Problems & 152 & 130 & 107 & 107 & 107 \\
\bottomrule
\end{tabular}
\end{table}

The results in Table~\ref{tab:perturbation_impact} show that SCSF's performance is strongly correlated with dataset similarity. As problems become more similar (perturbation size decreases), the speedup increases dramatically. The experiment also highlights the effectiveness of our sorting algorithm; SCSF consistently outperforms SCSF without sorting (`SCSF w/o sort`) across various similarity levels. In the theoretical limit of identical problems (0\% perturbation), the solution is found in just a few iterations. Conversely, for completely independent problems, SCSF's performance gracefully degrades to that of ChFSI, as expected.

\subsection{Analysis of Failure Cases: Discontinuous Datasets}
\label{sec:appendix_failure_case}

\rebuttal{
The core assumption of SCSF is that the dataset is generated from a process with underlying continuity, allowing our sorting algorithm to group similar problems effectively. 
To investigate the behavior of SCSF when this assumption is violated, We simulated a gradual mixing of the two datasets (Helmholtz and Poisson). We varied the proportion of Helmholtz operators from 100\% (a fully continuous dataset) down to 0\% (a different fully continuous dataset), with mixed ratios in between representing varying degrees of discontinuity. 
(dimension $6400$, $L=200$ eigenvalues) and solved them sequentially.
}

\begin{table}[h!]
\centering
\caption{Performance on a discontinuous dataset created by mixing Helmholtz and Poisson problems. All times are in seconds.}
\label{tab:discontinuous_case}
\begin{tabular}{@{}lccccc@{}}
\toprule
{Helmholtz \%} & {Eigsh} & {LOBPCG} & {ChFSI} & {SCSF (w/o sort)} & {SCSF} \\
\midrule
100\% (Helmholtz) & 152 & 130 & 107 & 46 & {31} \\
75\% & 154 & 203 & 116 & 94 & {78} \\
50\% (1:1 mix) & 154 & 280 & 132 & 118 & {98} \\
25\% & 152 & 359 & 141 & 108 & {80} \\
0\% (Poisson) & 149 & 454 & 153 & 81 & {52} \\
\bottomrule
\end{tabular}
\end{table}

\rebuttal{
Table~\ref{tab:discontinuous_case} presents the results. As expected, the performance gain of SCSF is reduced in this discontinuous scenario because the inter-problem correlation is disrupted, diminishing the effectiveness of the sorting module. However, even in this challenging case, SCSF still provides a notable speedup over baseline solvers, demonstrating a degree of robustness. The performance of `SCSF' approaches that of `SCSF (w/o sort)’, confirming that the sorting component's benefit is tied to dataset continuity.
}

\subsection{Cost-Benefit Analysis of the Sorting Algorithm}
\label{sec:appendix_sorting_overhead}

To address the trade-off between the cost of sorting and its benefits, we analyzed its computational overhead. Our analysis shows that the cost of the Truncated FFT Sort is negligible in the context of large-scale dataset generation. For example, as shown in Table~\ref{tab:sort2} of the main paper, sorting a dataset of $10^4$ samples takes approximately 151 seconds. In contrast, solving a single eigenvalue problem from the Helmholtz dataset can take over 250 seconds with a standard solver like Eigsh. For a full dataset of this size, the sorting overhead constitutes less than 0.1\% of the total generation time.

The benefit, however, is substantial. As shown in Table~\ref{tab:sort1}, sorting reduces the number of solver iterations by 5-50\% and total floating-point operations (Flops) by 7-43\%. Furthermore, our Truncated FFT Sort is significantly more cost-effective than a naive greedy sort, achieving nearly identical final solver performance at a fraction of the computational cost (see Tables~\ref{tab:sort2} and \ref{tab:sort3}). Given this highly favorable cost-benefit ratio, the sorting step is a crucial and efficient component of the SCSF framework.

\subsection{\rebuttal{Sensitivity to Parameterization}}
\label{app:sensitivity}

\rebuttal{
To assess the sensitivity of the proposed parameter sorting to different PDE parameterizations, we conducted an additional experiment comparing two distinct discretization methods for the Helmholtz problem: the {Finite Difference Method (FDM)} (utilized in the main text) and the {Finite Element Method (FEM/Galerkin)}.
}

\rebuttal{
As shown in Table \ref{tab:parameterization_sensitivity}, SCSF demonstrates consistent speedups across both discretization schemes. For the FEM dataset, we utilized a dimension of 10,000 with a tolerance of 1e-8.
}

\begin{table}[h]
\centering
\caption{Comparison of SCSF performance under different parameterizations (Helmholtz). Times are in seconds. `-' indicates non-convergence or excessive time cost.}
\label{tab:parameterization_sensitivity}
\begin{tabular}{lc|cccccc}
\toprule
Dataset & $L$ & Eigsh & LOBPCG & KS & JD & ChFSI & \textbf{SCSF (ours)} \\
\midrule
\multirow{2}{*}{{FDM (Central Diff)}} & 200 & 151.7 & 129.9 & 98.34 & 489.6 & 107.1 & {31.31} \\
 & 400 & 253.5 & 460.4 & 283.0 & 3829 & 121.5 & {40.52} \\
Dim: 6400, Tol: 1e-8 & 600 & 398.8 & 1031 & 329.6 & - & 146.2 & {51.32} \\
\midrule
\multirow{2}{*}{{FEM (Galerkin)}} & 200 & 623.1 & 515.7 & 365.6 & 1885 & 450.9 & {116.7} \\
 & 400 & 938.5 & 1746 & 1166 & - & 481.3 & {142.9} \\
Dim: 10000, Tol: 1e-8& 600 & 1499 & - & 1299 & - & 599.9 & {196.3} \\
\bottomrule
\end{tabular}
\end{table}

\rebuttal{
These results confirm that SCSF remains highly effective regardless of the discretization scheme, provided the relationship between parameters and operators remains continuous. 
}

\subsection{\rebuttal{Generalization of Truncation Level ($p_0$)}}
\label{app:p0_generalization}

\rebuttal{
To justify this universal constant, we calculated the ratio of the Frobenius norm of high-frequency components (frequency $> 20$) to the total norm for each dataset. The results are presented in Table \ref{tab:high_freq_ratio}.
}

\begin{table}[h]
\centering
\caption{Ratio of high-frequency component energy (frequency $> p_0=20$) to total energy across different PDE datasets.}
\label{tab:high_freq_ratio}
\begin{tabular}{lcccc}
\toprule
Dataset & Poisson & Ellipse & Helmholtz & Vibration \\
\midrule
High-frequency Ratio ($>p_0$) & 3.4\% & 3.7\% & 4.4\% & 4.8\% \\
\bottomrule
\end{tabular}
\end{table}

\rebuttal{
As shown, high-frequency components account for less than 5\% of the total energy across all diverse PDE families. This confirms that $p_0=20$ consistently captures over 95\% of the structural information inherent in the parameter matrices. 
}

\newpage

\section{\rebuttal{Theoretical Analysis of Truncated FFT Sorting}}
\label{ap_fft_theory}
\rebuttal{
Our sorting algorithm approximates the distance between operators to optimize the solving sequence. To analyze the theoretical validity of this approximation, we consider the isometric property of the Discrete Fourier Transform (DFT). Let $P^{(i)}$ and $P^{(j)}$ denote the discretized parameter matrices (e.g., diffusion coefficients) for two distinct problems. By Parseval’s identity, the DFT is a unitary transformation (up to a scaling factor), implying that the Frobenius distance in the spatial domain is strictly equivalent to that in the frequency domain:
}
$$
\| P^{(i)} - P^{(j)} \|_F^2 = \| \text{FFT}(P^{(i)}) - \text{FFT}(P^{(j)}) \|_F^2,
$$
\rebuttal{
where $ \text{FFT}(\cdot)$ denotes the normalized DFT. 
The SCSF algorithm computes a proxy distance using a truncated spectrum, retaining only frequencies within a low-frequency bandwidth $p_0$. Consequently, the exact distance can be decomposed into the sorting metric and a residual truncation error:
}
$$
\| P^{(i)} - P^{(j)} \|_F^2 = \underbrace{\| \text{Trunc}_{p_0}(\Delta \hat{P}) \|_F^2}_{\text{SCSF Metric}} + \underbrace{\| (I - \text{Trunc}_{p_0})(\Delta \hat{P}) \|_F^2}_{\text{Truncation Error } \epsilon_{ij}},
$$
\rebuttal{
where $\Delta \hat{P} = \text{FFT}(P^{(i)} - P^{(j)})$ and $ \text{Trunc}_{p_0}$ is the truncation operator.
}

\rebuttal{
The magnitude of the error term $\epsilon_{ij}$ is governed by the spectral decay properties of the underlying physical operators. The parameter matrices in our dataset, generated via Gaussian Random Fields (GRF) or representing physical media, typically exhibit smoothness or piecewise smoothness corresponding to Sobolev regularity $H^s$ with $s > 0$. Harmonic analysis dictates that the spectral energy of such functions decays polynomially or exponentially with frequency magnitude $|\mathbf{k}|$. Specifically, the Fourier coefficients satisfy $|\hat{P}_{\mathbf{k}}| \lesssim |\mathbf{k}|^{-s}$. Therefore, the truncation error is bounded by the tail energy of the spectrum:
$$
\epsilon_{ij} = \sum_{|\mathbf{k}| > p_0} |\Delta \hat{P}_{\mathbf{k}}|^2 \approx \mathcal{O}(p_0^{-2s+d}),
$$
where $d$ is the spatial dimension. This decay rate implies that the high-frequency components contribute negligibly to the global topological distance between operators. 
This is also the fundamental reason why truncated FFT sorting is effective.
}
% By filtering these components, the truncated FFT sorting not only reduces computational complexity from $\mathcal{O}(N^2)$ to $\mathcal{O}(p_0^2)$ but also acts as a low-rank approximation that effectively captures the dominant structural similarities required for efficient subspace recycling, while suppressing high-frequency numerical noise.

\end{document}

%% file: math_commands.tex
\usepackage{amsmath,amsfonts,bm}

\def\eqref#1{equation~\ref{#1}}
\def\1{\bm{1}}

\DeclareMathAlphabet{\mathsfit}{\encodingdefault}{\sfdefault}{m}{sl}
\SetMathAlphabet{\mathsfit}{bold}{\encodingdefault}{\sfdefault}{bx}{n}